# Rain Removal in Traffic Surveillance: Does it Matter?

Chris H. Bahnsen and Thomas B. Moeslund

***Abstract*— Varying weather conditions, including rainfall and snowfall, are generally regarded as a challenge for computer vision algorithms. One proposed solution to the challenges induced by rain and snowfall is to artificially remove the rain from images or video using rain removal algorithms. It is the promise of these algorithms that the rain-removed image frames will improve the performance of subsequent segmentation and tracking algorithms. However, rain removal algorithms are typically evaluated on their ability to remove synthetic rain on a small subset of images. Currently, their behavior is unknown on real-world videos when integrated with a typical computer vision pipeline. In this paper, we review the existing rain removal algorithms and propose a new dataset that consists of 22 traffic surveillance sequences under a broad variety of weather conditions that all include either rain or snowfall. We propose a new evaluation protocol that evaluates the rain removal algorithms on their ability to improve the performance of subsequent segmentation, instance segmentation, and feature tracking algorithms under rain and snow. If successful, the de-rained frames of a rain removal algorithm should improve segmentation performance and increase the number of accurately tracked features. The results show that a recent single-frame-based rain removal algorithm increases the segmentation performance by 19.7% on our proposed dataset, but it eventually decreases the feature tracking performance and showed mixed results with recent instance segmentation methods. However, the best video-based rain removal algorithm improves the feature tracking accuracy by 7.72%.**

## I. Introduction

MONITORING of road traffic is usually performed manually by human operators who observe multiple video streams simultaneously. However, the manual monitoring is both tiresome and does not scale with the growing number of cameras and an increased appetite for a deeper understanding of road user behavior. Thus, there is a clear-cut case for computer vision methods to step in and automate the process. If successful, vision methods in road user detection, classification, and tracking could give valuable insights into and analysis of road user behavior and accident causation, which could ultimately help reduce the number of accidents.

Manuscript received December 19, 2017; revised June 28, 2018; accepted September 15, 2018. This work was supported by Horizon 2020 through the European Union Framework Programme for Research and Innovation under Grant 635895. The Associate Editor for this paper was N. Zheng. *(Corresponding author: Chris H. Bahnsen.)*

The authors are with the Visual Analysis of People Laboratory, Aalborg University, 9000 Aalborg, Denmark (e-mail: cb@create.aau.dk).

However, most computer vision systems are designed to work under optimal conditions such as clear skies, low reflections, and few occlusions. Whenever one of these constraints is violated, the performance of the vision system rapidly deteriorates, and so does the promise of automated traffic analysis.

Non-optimal conditions are caused by several phenomena, but most prominently by bad weather conditions. We may divide bad weather into two main groups: steady and dynamic conditions [1]. Steady weather conditions include fog, mist, and haze, which degrade the contrast and reduce the visibility of the scene. Dynamic weather conditions include rainfall and snowfall, which appear as spatio-temporal streaks in the surveillance video, which may temporarily occlude objects with close proximity to the camera. Objects at greater distance from the camera are affected by the accumulation of rain and snow streaks, which reduces the visibility of the scene much like fog, mist, and haze.

We differentiate between three different approaches to cope with the challenges of bad weather in automated video surveillance: to mitigate the effects by pre-processing the video, to strengthen the robustness of the core vision algorithms, or to augment the sensing system by the use of multiple multi-modal sensors. In this work, we will study the implications of pre-processing the input video signal by algorithms that mitigate the dynamic effects of rainfall and snowfall. Many authors of such rain or snow removal algorithms note that these algorithms could help improve the robustness of traditional vision methods such as segmentation, classification, and tracking. We will investigate this claim through quantitative analysis and hereby provide valuable insights into this field for the benefit of the entire research community.

Current evaluations of rain removal algorithms are based on short video sequences or a collection of still images, typically provided by the authors themselves. Quantitative results are obtained by removing rain on synthetic datasets, where rain streaks are overlaid on rain-free images [2]. Is is common to see indoor images with synthetic rain as part of training [3] and testing [4] of rain removal algorithms. The performance of rain removal algorithms is usually measured by calculating the Structural Similarity Index (SSIM) [5] and the Peak Signal-to-Noise-Ratio (PSNR) between the de-rained and the rain-free images. However, a good SSIM or PSNR score on a synthetic dataset does not necessarily translate into performance when the rain removal algorithms are used on real-world footage. Such evaluation is typically performed by inspection of a limited selection of real-world rainy images.

We are curious how rain removal algorithms will work on traffic surveillance video under real-world conditions that include rainfall and snowfall, and how they affect a traditional computer vision pipeline. In this work, we want to measure the effectiveness of a rain removal algorithm, not by using the raw properties of the produced rain-removed images but by using the performance of the subsequent segmentation, instance segmentation, and feature tracking algorithms that run on top of the rain-removed imagery. If effective, a rain removal algorithm should improve the performance of the subsequent algorithms.

Our contributions are the following:

1) We provide a comprehensive overview of rain removal algorithms, using both single-image and video-based algorithms.
2) We provide a new publicly available dataset of 22 real-world sequences from 7 urban intersections in various degrees of bad weather involving rain or snowfall. Each sequence has a duration of 4-5 minutes and is recorded with both a color camera and a thermal camera.
3) We use this dataset and the BadWeather training sequences of the Change Detection 2014 challenge [6] to assess the performance of classic segmentation methods and recent instance segmentation methods on the raw and rain-removed imagery. Furthermore, we use the forward-backward feature tracking accuracy to investigate if feature-based methods perform better under rain-removed imagery.
4) The entire evaluation protocol and our implementation of the rain removal algorithm of Garg and Nayar [1] is publicly available as open-source[1,2] to enable others to build upon our results.

The rest of the paper is organized as follows: Section II describes how rain and snowfall impair the visual surveillance footage in traffic scenes. Section III gives a comprehensive overview of rain removal algorithms and their general characteristics. Our new dataset is presented in Section IV. The evaluation protocol of selected rain removal algorithms on this dataset is presented in Section V, and the results hereof are treated in Section VI. Section VII concludes our work.

## II. The Impact of Rain and Snow

Bad weather, including rain and snow, is generally acknowledged as a challenge in computer vision [7], but little work has been undertaken to identify the severity of this problem. In this section, we will shed light on the impact of rain and snow in traffic surveillance, and how it might affect the vision systems that are built on top of the video streams.

Rainfall and snowfall will have a negative effect on the visibility of the scene due to the atmospheric scattering and absorption from raindrops and snowflakes. In the atmospheric sciences, the combined impact of scattering and absorption from a particle is called extinction [8]. For wavelengths in the visible and infrared range, the extinction from raindrops can be approximated as [9]:

$$\beta_{\text{ext}_{\text{rain}}} = A \cdot R^B \tag{1}$$

where $\beta_{\text{ext}_{\text{rain}}}$ is the rain extinction coefficient, $A$ and $B$ are model parameters, and $R$ is the rain rate in mm/hr.

The exact values of the parameters $A$ and $B$ vary according to the precipitation type. Shettle [9] reports five different models for rain. We leave out the oldest rain model and plot the remaining four in Figure 1a. The approximation used in Equation 1 builds on the notion that the physical size of the raindrops is much greater than the wavelength in consideration. For radiation of longer wavelengths, a wavelength-dependent correction must be added [9].

The extinction caused by falling snowflakes in the visible range uses a model similar to Equation 1. However, one must convert the snow depth to equivalent liquid water. For wet snow, 1 mm of snow corresponds to 5 mm of rain, whereas for dry snow, the conversion range is greater than 1 to 20. For our calculations, we use the 'rule-of-thumb' approximation by Shettle [9] with a ratio of 1 to 10. When converted, the model is defined as:

$$\beta_{\text{ext}_{\text{snow}}} = A(1/10)^B \cdot S^B \tag{2}$$

where $\beta_{\text{ext}_{\text{rain}}}$ is the snow extinction coefficient, $S$ is the rate of snow accumulation, and other parameters are defined in Equation 1.

In Figure 1b, we plot the six different sets of model parameters reported by Mason [10]. Depending on the chosen model, one can see that the extinction from snow is a half-magnitude greater than the equivalent amount of rainfall. Thus, the visibility of a scene is reduced more under snow than under rain.

The properties of rainfall as they are observed by a typical surveillance camera have been studied extensively by Garg and Nayar [1], who laid out the theoretical framework for the physical and practical implications of rain in vision. They provided a physical model of a falling raindrop and constructed an appearance model for the raindrop as viewed by a camera. We will summarize the relevant findings of this work below:

1) Raindrops are transparent, and most drops are less than 1 mm in size.
2) The motion of a raindrop can be modeled as a straight line.
3) A raindrop appears brighter than its background. The change in intensities caused by a falling rain streak is linearly related to the background intensities that are occluded by the rain streak.

These observations, especially the notion that raindrops are brighter than the background, have had a major impact on all subsequent works on video-based rain removal. We will revisit these observations in our survey of rain removal algorithms in Section III. The implications of rain and snow on visual surveillance are not limited to the characteristics of a raindrop alone. The accumulation of rain on surfaces eventually leads to puddles when the drainage of the road is insufficient. When vehicles or other road users drive through these puddles, water will splash from the wheels. Raindrops may also attach to the lens or even freeze to ice if the camera is not installed inside a protective outdoor housing or the wind is too strong.

[1] https://bitbucket.org/aauvap/aau-rainsnow-eval
[2] https://bitbucket.org/aauvap/rainremoval/

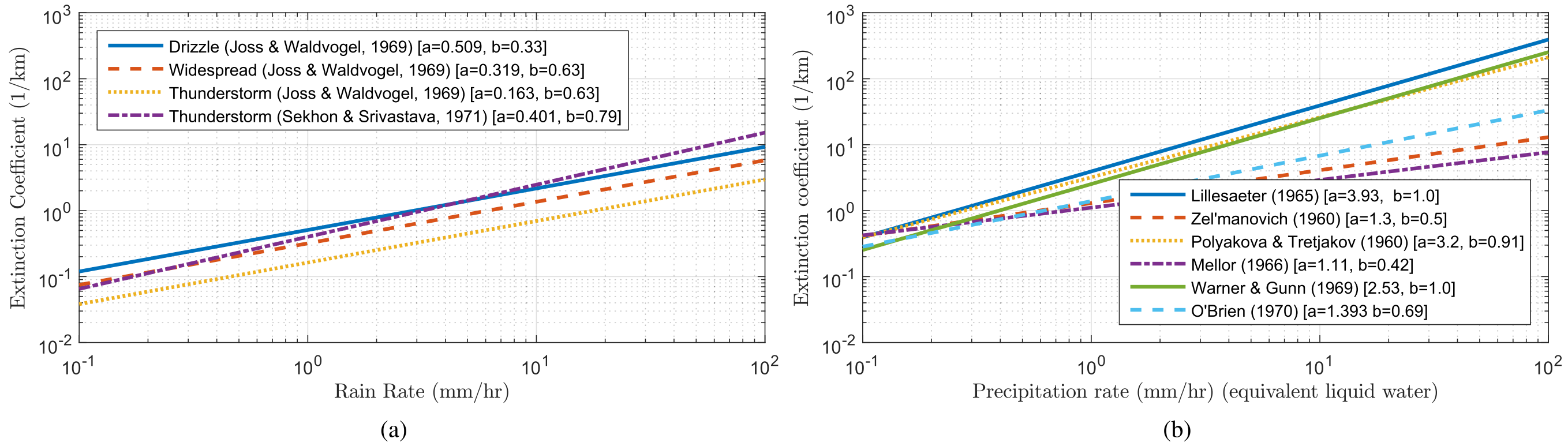

Fig. 1. Degradation of visibility due to extinction from rainfall and snowfall. Parameters as reported in [9] and [10]. (a) Extinction during rainfall. (b) Extinction during snowfall.

TABLE I
THE VISUAL EFFECTS OF RAIN AND SNOW

| Phenomena | Visual effect |
|---|---|
| Raindrop | Spatio-temporal streaks, duration approx. 1 frame per streak |
| Snowflake | Spatio-temporal streaks, duration approx. 1 frame per streak |
| Dense rain and snow | Reduced visibility, depth of field |
| Raindrops on lens | Blur, diffuse scattering of light |
| Puddles | Surroundings reflected by puddles and splashes from road users |

TABLE II
THE EFFECTS OF RAIN AND SNOW ON SEGMENTATION AND TRACKING

| Phenomena | Segmentation | Tracking |
|---|---|---|
| Raindrop | Non-constant background, false detections | False features, disturbances in tracking |
| Snowflake | Non-constant background, false detections | False features, disturbances in tracking |
| Dense rain and snow | Missing detections of far-away objects | Fewer salient features of far-away objects |
| Raindrops on lens | Missed detections in area of raindrop | No features in area of raindrop |
| Puddles | False detections due to reflections | False features due to reflections |

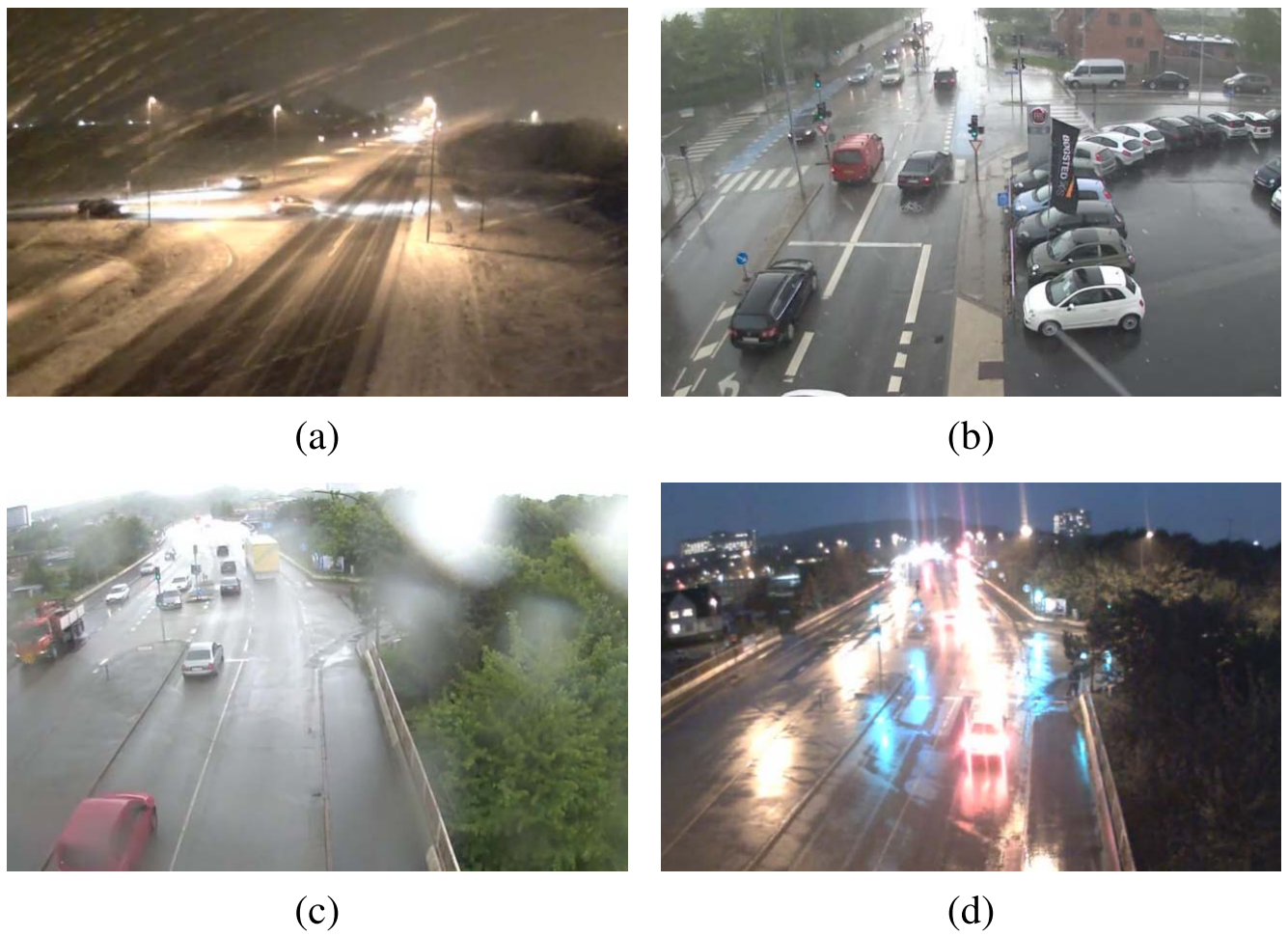

Fig. 2. Visual examples of rain and snow in traffic surveillance. In (a) and (d), bad lighting conditions further deteriorate the visibility of the scene. (a) Snow. (b) Heavy rain. (c) Raindrops on the lens. (d) Reflections.

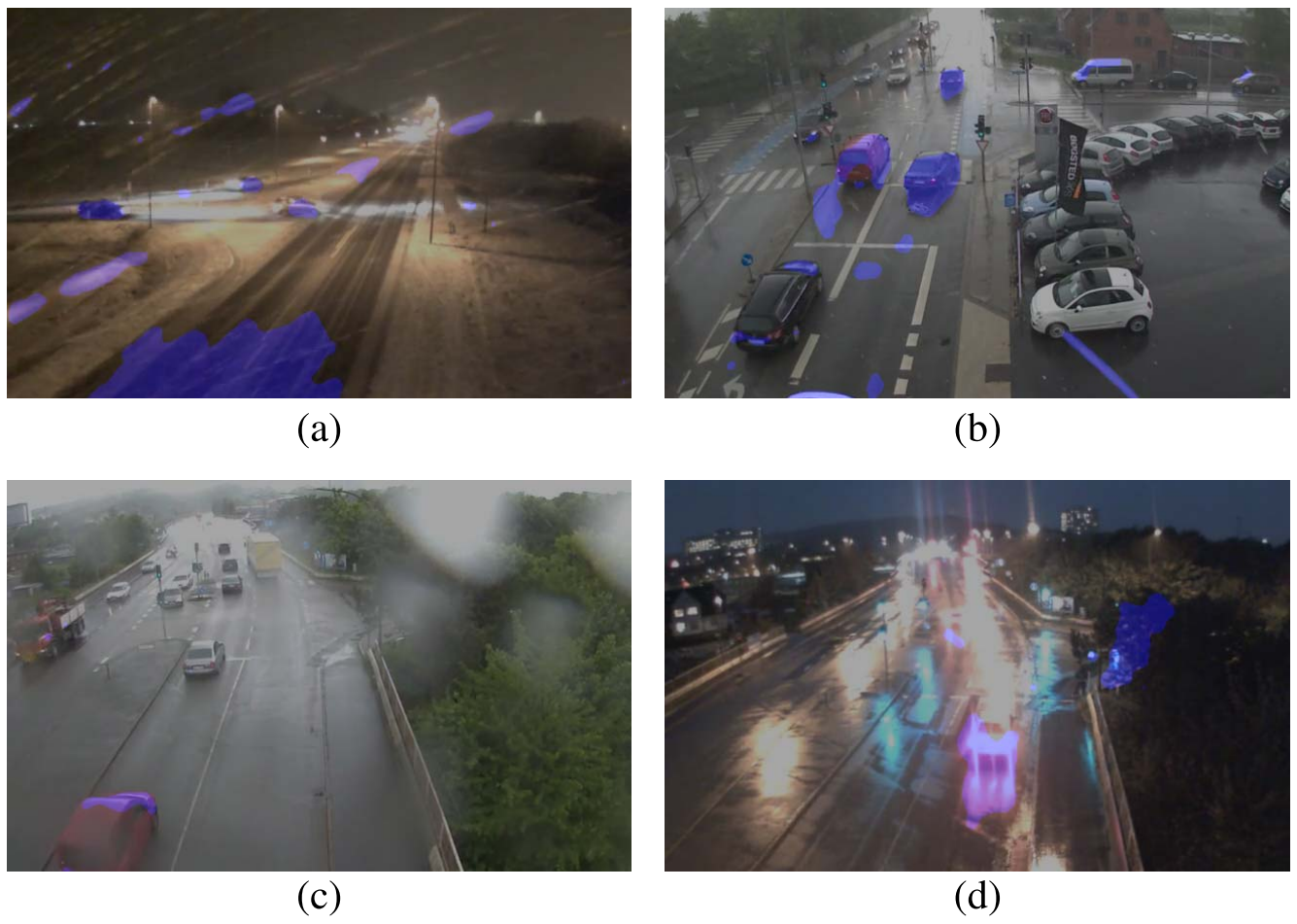

Fig. 3. Segmentation results of a state-of-the-art segmentation algorithm [11] on the sequences shown in Figure 2. The segmented masks are overlaid in blue. In (a) and (b), some snowflakes or rain streaks are detected as foreground. In (c), the raindrops on the lens lead to missing detections of parts of the red vehicle. In (d), reflections from the tail lights are detected as foreground, whereas most of the car is not. (a) Snow. (b) Heavy rain. (c) Raindrops on the lens. (d) Reflections.

Table I provides an overview of how these phenomena affect the observed images in a surveillance setting, whereas Figure 2 shows examples of footage impaired by snow, heavy rain, raindrops on the lens, and reflections on the road.

It is apparent from Figure 2 that these phenomena degrade the visibility of the road users. The degradation of visibility will inevitably affect vision algorithms due to a reduced signal-to-noise ratio. A detailed treatment on how vision-based segmentation and tracking algorithms are affected by the effects of rain and snow is given in Table II. The concrete effects on a state-of-the-art unsupervised segmentation algorithm [11] is shown in Figure 3.

We may also infer the impact of rain and snow from the results of existing challenges and datasets. The most prominent dataset on background segmentation, ChangeDetection.net [6], features a 'BadWeather' category that contains a total of 20,900 video frames distributed in four different scenes. Snow and snowfall are the common denominators for the 'BadWeather' sequences, but the exact nature of the scenes varies. In Table III, we have summarized the latest results of the ChangeDetection.net challenge for the BadWeather sequence, a trivial 'Baseline' sequence, and a weighted average of all sequences in the database. We distinguish between supervised change detection methods, which use the training samples of each sequence, and unsupervised change detection methods, which use default parameters for all sequences.

TABLE III
AVERAGE F-MEASURE OF THE BEST SEGMENTATION ALGORITHM ON THE CHANGEDETECTION.NET [6] DATABASE

| Method | BadWeather | Baseline | Overall |
|---|---|---|---|
| Best supervised | 0.98 | 0.99 | 0.97 |
| Best unsupervised | 0.87 | 0.96 | 0.79 |

The comparison between the BadWeather and Baseline sequences suggests that especially unsupervised change detection methods face difficulties when the video sequences are captured under bad weather conditions. However, the overall performance of the entire database is even lower than the BadWeather sequences. The best results on nighttime videos and the dynamic pan-tilt camera footage are dramatically lower with best f-measures of 0.76 for unsupervised change detection methods [6]. This suggests that a combination of rain or snow on nighttime videos may be increasingly difficult. It should be noted, however, that an extensive comparison of weather phenomena in surveillance requires that one can change only one parameter at the time, which is hardly the case with real-life surveillance footage.

## III. RAIN REMOVAL ALGORITHMS

The work within rain removal may be divided into two main categories: video-based rain removal, where the temporal information of a video stream is used to detect and remove the rain streaks, and single-image based rain removal, where such temporal information is not provided or used. We also make the distinction between rain streaks and rain drops. A rain streak is defined as a spatio-temporal effect with the approximate duration of one frame, whereas a rain drop is attached to the lens of the camera and remains stationary for seconds or even minutes. The published work on rain drop removal is significantly smaller than the corresponding work on rain streak removal. When temporal information is available, we consider rain drop removal to be considerably easier than rain streak removal. In this work, we will focus on the removal of rain streaks. However, the detection and removal of rain drops could be added as an additional pre-processing step, either before or after rain streak removal. Notable efforts on rain drop removal include the work of Eigen *et al.* [12], You *et al.* [13], and Roser and Geiger [14].

Rain removal can also be seen as a special case of image denoising. A good overview of general image denoising techniques is given in [15]. Image dehazing [16] and defogging [17] are also closely related fields that mitigate the effects of steady bad weather conditions.

An overview of rain streak removal techniques is available in Table IV. In the overview table, we note if the high-frequency (HF) parts of the image or video are explicitly computed as part of the rain streak removal. If so, we note the name of the used filter. Furthermore, we categorize the algorithms on the basis of how they learn from data and use the following criteria:

1) *Manual:* Requires manual hand-tuning of the algorithm for each particular sequence.
2) *Fixed:* The algorithm does not contain any adjustable parameters or parameters are provided 'as-is' by the original authors. Parameter tuning by the original authors is performed by hand and is based on empirical observations.
3) *Online:* One or more parameters are learned from the current input image or video. Contains no offline learning.
4) *Offline:* One or more parameters are learned from an offline database. The algorithm may also contain online learning.

The notion of snow removal is tightly coupled with the work within rain streak removal. In fact, the earliest rain removal technique of Table IV deals with noise elimination from snowfall [18]. Most authors of rain streak removal algorithms evaluate their rain removal algorithms solely on images of rainfall, but some also include images or videos of snowfall [19]–[21]. As summarized in Table I, the impact of rain and snow is similar in the visible spectrum, so it is natural to make a joint study of the two phenomenons. In the remainder of this article, we will refer to rain streak and snow removal jointly as rain removal.

### A. Single-Image Based Rain Removal

Rain removal from a single image is hard. Rain is a spatio-temporal phenomena and, without temporal information, one must make an informed guess on the temporal effects. Successful single-image based rain removal algorithms must effectively model the spatial influences of the rain streak and compensate accordingly.

We divide the published methods into four categories: filtering, matrix decomposition, dictionary learning, and convolutional neural networks.

*1) Filtering:* Applying one or more filters is a straightforward method to reduce the amount of rain in the image. A guided filter [56] is used in [44] to suppress the rain, where the minimum and maximum RGB values are used as the guidance image of the filter. In the work of Zheng *et al.* [4], the guided filter is used to split the image into low-frequency (LF) and high-frequency (HF) parts. The pixel-wise minimum of the LF image and of the input image is used as the input of an additional guided filter, which produces the rain-removed image. Although effective in suppressing the rain, the filter-based methods will also effectively blur textured and detailed parts of the image. In order to improve this, one needs to model the rain streaks.

TABLE IV

RAIN REMOVAL ALGORITHMS. WE LIST BOTH SINGLE-FRAME (IMAGE) AND VIDEO-BASED METHODS AND THEIR MAIN METHOD FOR RAIN STREAK DETECTION. MANUAL AND FIXED LEARNING INDICATE HAND-TUNING BY THE ORIGINAL AUTHORS, WHEREAS ONLINE AND OFFLINE LEARNING LEARNS FROM THE INPUT IMAGE, INPUT VIDEO, OR A COLLECTION OF OFFLINE IMAGES. IF THE METHOD COMPUTES HIGH-FREQUENCY IMAGES, WE NOTE THE NAME OF THE FILTER

| Paper | Year | Author group | Input | Color space | Main detection method | Learning | High-frequency image |
|---|---|---|---|---|---|---|---|
| [3] | 2017 | 1 | Image | RGB | Conditional generative adversarial network | Offline | No |
| [22] | 2017 | 1 | Image | RGB | Sparse dictionary | Offline | No |
| [23] | 2017 | 2 | Image | RGB | Convolutional neural network | Offline | Guided |
| [24] | 2017 | 2 | Image | RGB | Convolutional neural network | Offline | Guided |
| [25] | 2017 | None | Video | RGB | Matrix decomposition, orientation | Online | No |
| [26] | 2017 | None | Video | RGB | Matrix decomposition, intensity | Online | No |
| [2] | 2017 | None | Image | RGB | Convolutional neural network | Offline | No |
| [27] | 2017 | None | Image | RGB | Convolutional neural network | Offline | No |
| [28] | 2017 | None | Image | Grey | Sparse dictionary | Offline | No |
| [29] | 2017 | None | Image | RGB | Sparse dictionary, orientation, color | Offline | Guided |
| [30] | 2016 | None | Image | YUV | Matrix decomposition, dictionary | Offline | No |
| [20] | 2015 | 3 | Video | RGB | Sparse dictionary | Offline | No |
| [31] | 2015 | None | Image | RGB | Matrix decomposition, sparse dictionary | Online | No |
| [32] | 2015 | None | Video | RGB | Photometric, chromatic constraint, orientation | Fixed | No |
| [33] | 2014 | 4 | Image | RGB | Sparse dictionary | Online | Guided |
| [34] | 2014 | 4 | Image | RGB | Sparse dictionary | Online | Bilateral |
| [35] | 2014 | None | Image | RGB | Sparse dictionary | Online | Guided |
| [36] | 2014 | None | Image | HSV | Color thresholding, orientation | Fixed | No |
| [37] | 2014 | None | Video | RGB | Matrix decomposition | Online | No |
| [38] | 2014 | None | Video | RGB | Photometric constraint, optical flow | Fixed | No |
| [39] | 2013 | 3 | Image | RGB | Streak orientation, intensity | Fixed | No |
| [40] | 2013 | 5 | Video | YCbCr | Photometric constraint, orientation | Fixed | No |
| [41] | 2013 | None | Video | RGB | Matrix decomposition | Online | No |
| [4] | 2013 | None | Image | RGB | Filtering | Fixed | Guided |
| [42] | 2012 | 4 | Video | RGB | Sparse dictionary | Online | Bilateral |
| [43] | 2012 | 4 | Image | RGB | Sparse dictionary | Online | Bilateral |
| [44] | 2012 | 6 | Image | YUV | Filtering | Fixed | No |
| [45] | 2012 | None | Video | RGB | Photometric constraint | Fixed | Bilateral |
| [46] | 2012 | 5 | Video | YCbCr | Photometric constraint | Fixed | No |
| [47] | 2011 | 4 | Image | RGB | Sparse dictionary | Online | Bilateral |
| [48] | 2011 | 5 | Video | RGB | Photometric constraint, orientation | Fixed | No |
| [49] | 2011 | None | Video | RGB | Photometric constraint | Fixed | No |
| [21] | 2011 | None | Video | RGB | Background segmentation, photometric constraint, orientation | Online | No |
| [50] | 2011 | None | Video | RGB | Filtering, chromatic constraint, streak intensity | Fixed | No |
| [19] | 2010 | None | Video | RGB | Frequency space, streak orientation, intensity | Manual | No |
| [51] | 2009 | 6 | Video | RGB | Photometric, chromatic constraint | Fixed | No |
| [52] | 2008 | None | Video | RGB | Photometric constraint, streak orientation | Fixed | No |
| [53] | 2008 | None | Video | RGB | Photometric constraint, streak orientation | Fixed | No |
| [1] | 2007 | None | Video | RGB | Photometric constraint, streak orientation, size | Fixed | No |
| [54] | 2006 | None | Video | RGB | Streak intensity, chromatic constraint | Online | No |
| [55] | 2003 | None | Video | RGB | Temporal median | Fixed | No |
| [18] | 1999 | None | Video | RGB | Temporal median | Fixed | No |

*2) Matrix Decomposition:* A rain streak model is obtained in matrix decomposition techniques by adding additional constraints on the removal process. The rain removal problem is formalized as an exercise in decomposing the input image $I$ into the rain-free image $B$ and the rain image $R$, such that they add up to comprise the original image:

$$I = B + R \tag{3}$$

In order to guide the decomposition, one has to make certain assumptions on the properties of $B$ and $R$. A common assumption is that $B$ should have low total variation (TV), i.e. the recovered image should be smooth:

$$\mathrm{TV}_B = \sum_{i=1}^{M}\sum_{j=1}^{N} \|\Delta b_{i,j}\|_2 \tag{4}$$

where $M$ and $N$ are the dimensions of the image, and $\Delta b_{i,j}$ is the gradient of $B$ at position $i, j$.

With the exception of dense rain, rain streaks appear relatively infrequently compared to the total number of pixels in an image. Thus, it makes sense to impose sparsity on $R$ via the squared Frobenius norm [30]:

$$\arg\min_R = \|R\|_F^2 = \sum_{i=1}^{M}\sum_{j=1}^{N} |r_{i,j}|^2 \tag{5}$$

where $r_{i,j}$ is the pixel value of $R$ at position $i, j$.

Other methods utilize different norms to induce sparsity. In [22], the 1-norm is used to induce sparsity on $B$:

$$\arg\min_B = \|B\|_1 = \sum_{i=1}^{M}\sum_{j=1}^{N} |b_{i,j}| \tag{6}$$

where $b_{i,j}$ is the pixel value of $B$ at position $i, j$.

Some methods [26], [37] use the nuclear norm to enforce low rank on $B$:

$$\arg\min_{B} = \|B_{i,j}\|_* = \text{tr}(\sqrt{B^* B}) \quad (7)$$

*a) Solvers:* Different methods have been proposed to solve the constrained matrix decomposition problem, e.g. the Alternating Direction Method of Multipliers (ADMM) [57], (Robust) Principal Component Analysis (PCA), or the Inexact Augmented Lagrange Multiplier (IALM) [58]. Other solvers are typically used in combination with dictionary learning, which is described in the following. When the decomposition is converging, the background image $B$ is used as the rain-removed image.

The rain removal methods that use a matrix decomposition scheme based on Equation 3 or variants of it are listed in Table V, which shows that the community does not agree on the constraints for $R$ and $B$. Some works demand sparsity on $R$ [37], while others demand it on $B$ [22]. The same holds for the low rank requirement of $R$ and $B$. The disagreement sheds light on the general problem of matrix composition techniques: neither the rain nor the background is guaranteed to adhere to the imposed mathematical constraints on the image structure. Thus, there might be high-frequency textures 'trapped' within the segmented rain image and rain streaks 'trapped' within the segmented background image.

*3) Dictionary Learning:* Based on the observation that rain streaks fall in the same direction and share similar patterns, one can formulate the rain removal problem as segmenting the image into patches. These patches are classified into rain and non-rain patches by one or more dictionaries. The dictionaries may be learned online from the input image or from a offline bank of (generated) rain streaks.

Dictionaries were introduced in rain removal by Fu *et al.* [47] and improved by Kang *et al.* [43]. The authors applied a variant of the Morphological Component Analysis (MCA) technique [59] on the HF component of the input image to decompose the image into $B$ and $R$. A sparse coding algorithm [60] is used to learn a dictionary of atoms from patches of the HF image. The Histogram of Oriented Gradients (HOG) is computed for each atom, and the output hereof is used as input to a two-cluster K-means algorithm. The cluster with the smallest gradient variance is selected as the rain atoms. Once the dictionary is classified, Orthogonal Matching Pursuit (OMP) [61] is used to sparsely reconstruct the HF image:

$$\min_{\theta_{\text{HF}}^k} = \|b_{\text{HF}}^k - D_{\text{HF}} \cdot \theta_{\text{HF}}^k\|_2^2 \quad \text{s.t.} \ \|\theta_{\text{HF}}^k\|_0 \leq L \quad (8)$$

where $D_{\text{HF}}$ is the dictionary containing both rain and non-rain atoms of the HF image, $b_{\text{HF}}^k$ is the k'th patch of the HF image, $\theta_{\text{HF}}^k$ is a matrix containing the sparse coefficients for reconstructing the k'th patch, and $L$ is the maximum number of non-zero elements in $\alpha$. In [43], $L = 10$.

The dictionary components of $D_{\text{HF}}$, which corresponds to the previously classified non-rain atoms, are used to reconstruct the HF part of the rain-removed image. Finally, this image is added to the low frequency (LF) image, and a rain-removed image is obtained. In our experiments, we have experienced that the sparse reconstruction may be entirely composed of what is classified as rain atoms. In this case, the rain-removed image is completely empty.

Huang *et al.* [42] used the same rain removal framework as [43] but changed the selection of rain and non-rain atoms. Instead of using K-means on the HOG-computed gradient, they used Principal Component Analysis (PCA) to find the dominant directions of intensity change. The most similar and dissimilar atoms are assigned as rain and non-rain atoms, respectively. The remaining atoms are classified by a Support Vector Machine (SVM).

A different variant hereof is presented in [34]. The HOG- and PCA-based approaches are still used to find the discriminative features of the dictionary components, but the grouping of atoms is performed by the use of affinity propagation [62]. A greedy scheme is performed on top hereof to select $K <= 16$ clusters. The variance of the atoms in each cluster is computed, and the cluster with the lowest variance is regarded as the one containing the rain atoms. The remaining clusters are then used to reconstruct the rain-removed image.

Another augmentation of [43] is provided by Chen *et al.* [33], who introduced a depth of field measure on the atom components. Pixels with a low depth of field are regarded as rain. Furthermore, the work uses chromacity information of the atoms to restore details that might otherwise be regarded as rain atoms. In the independent work of Wang *et al.* [29], the authors use the color variance of an atom to refined the HF rain image from [43].

The main shortcoming of the above family of dictionary learning techniques [33], [42], [43], [47] is the unilateral dependence on the input image for both the rain and non-rain dictionaries. Even when augmented with additional constraints by Huang *et al.* [42] and Chen *et al.* [33], HF details may be integrated into the rain dictionaries and thus be removed from the rain-removed image. In order to combat this problem, it seems that one should resort to offline techniques for dictionary training.

A combination of offline and online training is used by Li *et al.* [30], who utilize Gaussian Mixture Models (GMM) to learn two dictionaries of rain and non-rain patterns. The non-rain dictionary is learned offline, while the rain streak dictionary is learned using relatively flat regions of the image. The generation of the background and rain image is formulated as an image decomposition problem:

$$\min_{B,R} \|I - B - R\|_F^2 + \alpha\|\Delta B\|_1 - \beta\|R\|_F^2 - \gamma G(B, R) \quad (9)$$

where $I$, $B$, and $R$ are the input, background, and rain image respectively. $\alpha$, $\beta$, and $\gamma$, are scalars estimated heuristically, and $G(B, R)$ is the reconstruction of the background and rain image based on the respective dictionaries.

One sees from Equation 9 that further constraints are put on the reconstructed background and rain image; the background must have low total variation as defined in Equation 4, and the rain image must be sparse as defined by the Frobenius norm. The decomposition problem is solved by the L-BFGS

TABLE V
RAIN REMOVAL ALGORITHMS THAT INCLUDE DECOMPOSITION OR DICTIONARY LEARNING

| Paper | Year | Author Group | Matrix decom-position | Dic-tionary learning | Assumptions | Solver or method | Patch-based | High-frequency image |
|---|---|---|---|---|---|---|---|---|
| [22] | 2017 | 1 | X | X | $R$ low rank, $B$ sparse, $B$ low TV | ADMM | X | No |
| [25] | 2017 | None | X | | $B$ low temporal variation, $B$ low TV on x, y-axes | ADMM | | No |
| [26] | 2017 | None | X | | $B$ low rank, $R$, $F$ sparse, $R_d$ Gaussian | MRF, SVD | | No |
| [28] | 2017 | None | | X | $R$ vertical gradients, $B$ high SSIM | OMP, Genetic algorithm | X | No |
| [29] | 2017 | None | | X | $R$ horizontal gradients, low color, gradient variance | HOG | X | Guided |
| [30] | 2016 | None | X | X | $R$ sparse, $B$ low TV, $B$, $R$ $n$-modal Gaussian | L-BFGS, GMM | X | No |
| [20] | 2015 | 3 | | X | $R$ similar angular components | MCA, SVD, SVM | X | No |
| [31] | 2015 | None | X | X | $R$, $B$ mutually exclusive | OMP, K-SVD | X | No |
| [33] | 2014 | 4 | | X | $R$ low depth of field, $B$ chromatic | MCA, HOG | X | Guided |
| [34] | 2014 | 4 | | X | $R$ similar gradients | MCA, HOG, PCA | X | Bilateral |
| [35] | 2014 | None | | X | $R$ high structural similarity | SSIM, K-SVD | X | Guided |
| [37] | 2014 | None | X | | $R$ sparse, $B$ low rank | PCA | | No |
| [41] | 2013 | None | X | | $R$ linearly dependent, $B$ low TV | IALM | X | No |
| [42] | 2012 | 4 | | X | $R$ similar gradients | MCA, HOG, PCA, SVM | X | Bilateral |
| [43] | 2012 | 4 | | X | $R$ low gradient variance | MCA, HOG, K-means | X | Bilateral |
| [47] | 2011 | 4 | | X | $R$ brighter than neighbors | MCA, K-SVD | X | Bilateral |

TABLE VI
RAIN REMOVAL ALGORITHMS THAT USE CONVOLUTIONAL NEURAL NETWORKS

| Paper | Year | Author group | Network type | Convolutional layers | Training size | High-frequency image |
|---|---|---|---|---|---|---|
| [3] | 2017 | 1 | Conditional Generative Adversarial Network (Pix2Pix [65]) | 12 (Generator) | 700 | No |
| [23] | 2017 | 2 | Convolutional Neural Network | 3 | 4900 | Guided |
| [24] | 2017 | 2 | Convolutional Neural Network (ResNet [64]) | 26 | 9100 | Guided |
| [2] | 2017 | None | Convolutional Neural Network (Dilated convolution) | 10 | $< 300$ | No |
| [27] | 2017 | None | Convolutional Neural Network (Inception-V4 [66]) | $> 76$ | 16.000 | No |

algorithm [63]. The rain removal algorithms that use dictionary learning are listed in Table V.

*4) Convolutional Neural Networks:* Convolutional Neural Networks (CNNs) were introduced to rain streak removal almost simultaneously in [2], [3], [23], and [24]. We summarize the work on CNN-based rain removal in Table VI.

A classical CNN approach was proposed by Fu *et al.* [23]. Just like prior work in dictionary learning, the rain removal is performed on the HF components of the input image, which is produced by the guided filter. The network uses three convolutional layers to de-rain the HF image, which is subsequently added to the LF image. Training is performed on synthesized rain images using rain streaks generated in Adobe Photoshop. Subsequent work by the same author [24] used a much deeper network with residual connections [64] and an increased number of training samples (9100).

Yang *et al.* [2] used dilated convolutions on three different scales [67] to aggregate multi-scale information due to the variable-size receptive field of the dilated convolutions. This helps the network to incorporate contextual information, which might help when learning to remove the rain.

Inspired by the success of generative networks, Zhang *et al.* [3] used a generative adversarial network (GAN) that is conditioned on the input image. They used the Pix2Pix framework [65] to create a generator network that de-rains an input image whereto artificial rain streaks have been added. Based on appearance, it is the role of an additional discriminator network to judge whether a de-rained image is the output of the generator network or is the original rain-free image.

A dedicated CNN-framework for snow removal is proposed by Liu *et al.* [27]. The snow removal network consists of a translucency and residual recovery module that handles the restoration of the snow-free image from semi-transparent and fully opaque snow streaks, respectively. The architecture is inherited from Inception-v4 [66] and enhanced by using the atrous spatial pyramid pooling from DeepLab [68].

## B. Video-Based Rain Removal

The first attempts of removing rain in video sequences took advantage of the short duration of a rain streak, i.e. that a single streak is visible to the camera in one frame and then disappears. This means that the rain removal problem may be formalized as a low-pass filtering problem in which the rain streaks are unwanted high-frequency fluctuations. As such, the rain will be removed by applying a temporal median filter on the entire image [18], [55]. The problem with this approach is that all other temporal motion will be blurred too.

*1) Photometric Constraint:* In order to prevent blurring of the non-rain image, it is therefore beneficial to detect the individual rain streaks. Such detection was introduced by Garg and Nayar [69] when they studied the photometry of falling raindrops. As we described in Section II, they introduced a method to find candidate rain pixels based on the observation that a raindrop appears brighter than the background and that each rain streak only appears in a single frame. Thus, under the assumption that the background

TABLE VII

RAIN REMOVAL ALGORITHMS THAT DO NOT INCLUDE DECOMPOSITION, DICTIONARY LEARNING, OR NEURAL NETWORKS. I: IMAGE, V: VIDEO

| Paper | Year | Detection method | | | | Removal method |
|---|---|---|---|---|---|---|
| | | Input | Photometric constraint | Chromatic constraint | Streak orientation | |
| [32] | 2015 | V | X | X | X | Temporal blending |
| [36] | 2014 | I | | X | X | Inpainting |
| [38] | 2014 | V | X | X | | Spatio-temporal mean |
| [39] | 2013 | I | | | X | Nonlocal means filter |
| [40] | 2013 | V | X | | X | Temporal mean |
| [4] | 2013 | I | | | | Guided filter |
| [44] | 2012 | I | | | | Guided filter |
| [45] | 2012 | V | X | | | Inpainting |
| [46] | 2012 | V | X | | | Temporal mean |
| [48] | 2011 | V | X | | X | Temporal mean |
| [49] | 2011 | V | X | | | Spatio-temporal mean |
| [21] | 2011 | V | X | | X | None |
| [50] | 2011 | V | | X | X | Blending |
| [19] | 2010 | V | | | X | Temporal mean |
| [51] | 2009 | V | X | X | | Kalman filter |
| [52] | 2008 | V | X | | X | Temporal mean |
| [53] | 2008 | V | X | | X | Temporal mean |
| [1] | 2007 | V | X | | X | Temporal mean |
| [54] | 2006 | V | | X | X | Temporal blended mean |
| [55] | 2003 | V | | | | Temporal median |
| [18] | 1999 | V | | | | Temporal median |

remains stationary, the candidate pixels that may contain a rain streak should satisfy the following condition:

$$\Delta I = I_n - I_{n-1} = I_n + I_{n+1} \geq c \tag{10}$$

where $I_n$ denotes the image at frame $n$, and $c$ is the minimum intensity change to distinguish rain drops, fixed to $c = 3$. For each frame, the candidate streaks are refined by the requirement that the intensity change of pixels on the same streak should be linearly related to the background intensities, $I_b$, at time $n-1$ and $n+1$:

$$\Delta I = -\beta I_b + \alpha \tag{11}$$

This should hold for all pixels within a single streak as imaged by a single frame if $\beta$ is within the range [0; 0.039] [69]. The step performed in Equation 11 is denoted as the photometric constraint. However, in subsequent work on video-based rain removal, also the constraint of Equation 10 is denoted as the photometric constraint. In our overview of rain removal methods in Table IV, we use the term 'photometric constraint' if either Equation 10 or 11 have been applied. In the work by Garg and Nayar, the binary output of the photometric constraint is correlated for a temporal window of 30 frames. Spatio-temporal streaks that have a strong directional component are regarded as the detected rain streaks. In Table IV, we refer to this, and variants hereof, as the streak orientation constraint. Streaks that consist of only a few pixels will be filtered out during this selection as their Binary Large OBjects (BLOBs) will not impose a strict directional structure.

The detected rain streaks are removed by using the two-frame average of frame $n-1$ and $n+1$, i.e. the temporal mean. Rain removal algorithms that use the photometric constraint, for example, are summarized in Table VII. These algorithms typically include a separate detection and removal step, where detected rain pixels are smoothed out by using either a temporal or spatial filter. We denote this as the 'removal method' in Table VII.

*2) Chromatic Constraint:* The intensity-based temporal constraint of Equation 10 is usually applied on gray-scale images. In [54], the constraint is extended to color images by assuming that the temporal differences of the three color channels are approximately similar when the background is occluded by a rain streak, otherwise not.

*3) Streak Orientation:* A background subtraction algorithm is used in [21] to generate candidate streaks, which are refined by the selection rule of Equation 10 and the removal of large BLOBs. The orientation of the remaining streak candidates is modeled by a Gaussian-uniform mixture distribution. By the assumption of the similar orientation of rain streaks, rain streaks are detected if the Gaussian part of the mixture distribution is dominant relative to the uniform part.

Barnum *et al.* [19] analyzed the properties of rain streaks in frequency space and found that the rain streaks impose a strong directional component in the Fourier-transformed image. By thresholding the rotation and magnitude of the Fourier-transformed videos, they are capable of detecting most of the rain. As the rotation of rain streaks is dependent on the wind, one has to manually tune the ratios for each rainfall.

*4) Matrix Decomposition:* The intensity fluctuations with respect to a background model are used as an initial estimate of sparse rain streaks and the foreground in [26]. These are used as the initial estimates of a matrix decomposition problem, where the image is decomposed into background, foreground, dense streaks, and sparse streaks:

$$I = B + F + R_s + R_d \tag{12}$$

where $I$ is the input image, $B$ is the background, $F$ is the foreground, and $R_s$, $R_d$ are the sparse and dense rain streaks, respectively.

The decomposition is enabled by a Markov Random Field (MRF), which uses optical flow from adjacent frames to detect moving objects from which rain removal is performed by using similar patches in adjacent frames.

Jiang *et al.* [25] expanded the matrix decomposition problem into a tensor decomposition problem by integrating the adjacent frames in the decomposition. They assumed that rain streaks are vertical and thus proposed to minimize the $l_0$ and $l_1$ norm of the total variation on the x and y axes, respectively. It is, furthermore, assumed that the temporal difference between the rain-removed frames is minimal. These constraints are solved using the Alternating Direction Method of Multipliers (ADMM) [57].

*5) Dictionary Learning:* Instead of using the candidate pixel selection of Equation 10, Kim *et al.* [20] used two-frame optical flow to generate the frame difference, which is used as the initial rain map. The rain map is decomposed into sparse basis vectors corresponding to patches of size $16 \times 16$ pixels. A pre-trained SVM classier is used to filter the rain streaks from noise based on the orientation of the patch. The rain-removed image is restored using rain-free patches from adjacent frames in an Expectation-Maximization (EM) scheme.

### C. Rain Removal Benchmarks

As mentioned in the introduction, existing evaluations of rain removal algorithms are based on short video sequences or a collection of images from the authors. Quantitative evaluation is typically performed on a set of rain-free images, where synthetic rain is overlaid. The synthetic rain is either produced in Adobe Photoshop[3] [23], reused from the work of Garg and Nayar [70], which considers the photo-realistic rendering of single rain streaks, or produced by using own methods [19], [71]. For videos, the synthetic rain is produced in Adobe After Effects [20].

The two most popular metrics for comparing the rain-removed image and the original rain-free image are the Structural Similarity Index (SSIM) [5] and the Peak Signal-to-Noise-Ratio (PSNR). Other common metrics include the Visual Information Fidelity (VIF) [72] and the Blind Image Quality Index (BIQI) [73]. BIQI is a no-reference algorithm for assessing the quality of an image, meaning that it does not require the rain-free image for comparison. Other metrics include the forward-backward feature tracking accuracy [19], the measurement of image variance [46], and the comparison of face detection scores on original and rain-removed images [32]. An overview of how the existing rain removal algorithms are evaluated is listed in Table VIII. It should be noted that we have only included comparisons with dedicated rain removal algorithms and thus excluded comparisons with general-purpose noise removal or image filtering algorithms.

It is observed from Table VIII that only a few competing rain removal algorithms are evaluated for each proposed method, thus hindering a general comparison of the entire field. Fortunately, recent works on rain removal include a more thorough evaluation on competing algorithms. On average, the rain removal algorithms published in 2017 have been evaluated on approximately three competing algorithms. However, a true overview of the performance across algorithms remains a challenge. This is caused by the following:

1) Few authors have made their implementations publicly available.
2) There is limited availability of public datasets for validation.

The implementations of [3], [20], [24], [34], [42], and [43] are available to the general public. If one wants to compare other methods, they must be re-implemented manually, which does not guarantee comparable performance nor comparable results.

A few public datasets have recently emerged. Li *et al.* [30] introduced a dataset with 12 images,[4] all with and without artificial rain. Along with their open-source implementation of their proposed rain removal algorithm, Zhang *et al.* [3] also made their training and test sets available; these consist of a total of 800 images.[5] For video-based rain removal, unfortunately, no such dataset exists. Thus, any application-based evaluation of rain removal algorithms are hindered due to lack of appropriate datasets.

TABLE VIII

EXISTING EVALUATIONS OF RAIN REMOVAL ALGORITHMS. I: IMAGE, V: VIDEO

| Paper | Year | Comparison metrics | | | | | Compared methods |
|---|---|---|---|---|---|---|---|
| | | Input | SSIM | PSNR | VIF | Other | |
| [3] | 2017 | I | X | X | X | X | [22], [23], [31], [41], [43] |
| [22] | 2017 | I | X | X | | | [23], [31], [41], [43] |
| [23] | 2017 | I | X | | | X | [30], [31], [34] |
| [24] | 2017 | I | X | | | | [30], [31] |
| [25] | 2017 | V | X | X | | X | [20], [30], [31] |
| [26] | 2017 | V | | X | | | [1], [20], [46], [54] |
| [2] | 2017 | I | X | X | | | [30], [31], [43] |
| [27] | 2017 | I | X | X | | | [23] |
| [28] | 2017 | I | X | X | X | X | [34], [43] |
| [29] | 2017 | I | X | X | | X | [30], [31], [33] |
| [30] | 2016 | I | X | | | | [41], [43] |
| [20] | 2015 | V | | X | | | [1], [19], [43], [54] |
| [31] | 2015 | I | X | X | | | [20], [43] |
| [32] | 2015 | V | | | | X | [1], [54] |
| [33] | 2014 | I | | | X | | [1], [43] |
| [34] | 2014 | I | | | | | |
| [35] | 2014 | I | X | X | | | [34] |
| [36] | 2014 | I | | | | | [43] |
| [37] | 2014 | V | | | | | [1] |
| [38] | 2014 | V | | | | | [1], [46], [54] |
| [39] | 2013 | I | | | | | [43] |
| [40] | 2013 | V | | | | X | [1], [19], [46], [48], [51] |
| [41] | 2013 | V | | | X | | [19], [43], [44] |
| [4] | 2013 | I | | | | | [44] |
| [42] | 2012 | V | | X | | | [43] |
| [43] | 2012 | I | | | X | | |
| [44] | 2012 | I | | | | | |
| [45] | 2012 | V | | | | | [1] |
| [46] | 2012 | V | | | | X | [1], [48], [51], [54] |
| [47] | 2011 | I | | | | | |
| [48] | 2011 | V | | | | X | [1], [51], [54] |
| [49] | 2011 | V | | | | | [1] |
| [21] | 2011 | V | | | | | |
| [50] | 2011 | V | | | | | |
| [19] | 2010 | V | | | | X | [1], [54] |
| [51] | 2009 | V | | | | | [1], [54] |
| [52] | 2008 | V | | | | | [1] |
| [53] | 2008 | V | | | | | [1], [54] |
| [1] | 2007 | V | | | | | |
| [54] | 2006 | V | | | | | [1] |
| [55] | 2003 | V | | | | | |
| [18] | 1999 | V | | | | | |

### D. Common Challenges of Rain Removal Algorithms

In Table IX, we summarize the underlying assumptions and the main challenges of the reviewed rain removal algorithms. It is interesting to note that even the sophisticated algorithmic methods of matrix decomposition and sparse dictionary are governed by heuristic assumptions that not necessarily translate to real-world conditions. The recent advent of CNNs in rain removal is promising but relies on a collection of synthetic images for training. The generation of more realistic, synthetic rain as well as the introduction of synthetic rain in longer video sequences could help move the frontiers in image and especially video-based rain removal.

## IV. NEW DATASET

In order to thoroughly evaluate the performance of rain removal algorithms under real-world conditions, we introduce the AAU RainSnow dataset[6] that includes 22 challenging sequences captured from traffic intersections in the Danish cities of Aalborg and Viborg. The sequences of the dataset are captured from seven different locations with both a conventional RGB camera and a thermal camera, each with a resolution of 640 x 480 pixels at 20 frames per second. We provide color information for the conventional RGB camera as some rain removal algorithms are explicitly created for color images and some segmentation algorithms produce better results with color than gray-scale images [74].

[3]http://www.photoshopessentials.com/photo-effects/rain/

[4]http://yu-li.github.io/

[5]https://github.com/hezhangsprinter/ID-CGAN/

[6]https://www.kaggle.com/aalborguniversity/aau-rainsnow

TABLE IX
COMMON CHALLENGES OF RAIN REMOVAL ALGORITHMS

| Detection method | Image | Video | Assumptions | Problems |
|---|---|---|---|---|
| Spatial filtering | X | | $R$ high-frequency patterns | Textured objects detected as $R$ |
| Temporal filtering | | X | Static $B$, moving $R$ | Moving objects detected as $R$ |
| Photometric, chromatic constraint | | X | Semi-static $B$, moving $R$ | Moving light colored objects detected as $R$ |
| Streak orientation | X | X | $R$ fall in a particular direction | $R$ orientation vary according to the wind |
| Matrix decomposition | X | X | $R$ patches share unique descriptors | No guarantee that descriptors adhere to real world |
| Sparse dictionary | X | | $R$ patches share unique patterns | Some $R$ patterns are shared with $B$ |
| CNN | X | X | $R$ patterns can be learned offline | Requires synthetic images for training |

Rain and snow are the common denominators of the sequences. In some sequences, the rain is very light and mostly visible as temporal noise. In other sequences, the rain streaks are clearly visible spatial objects. The illumination conditions vary from broad daylight to twilight and night. When combined with rain, snow, moist, and occasional puddles on the road, the variations in lighting create several challenging conditions, such as reflections, raindrops on the camera lens, and glare from headlights of oncoming cars at night.

The characteristics of each scene in our dataset are listed in Table X. The weather conditions and the temperature for each scene have been estimated by correlating the observed weather with publicly available weather station data.[7] The distance from the weather station to the scene is 25 km for the Ringvej sequences and a maximum of 13 km for all other sequences. We also include key characteristics of the BadWeather sequences in the ChangeDetection.net dataset, as shown in Table X, to enable a comparison of the two datasets.

One observes from Table X that our dataset comprises of more objects per frame and that the observed objects are significantly smaller than the BadWeather sequences. Falling snow is present in all of the BadWeather sequences, but the lighting conditions are fine, with all areas of the scene being sufficiently lit. Thus, we believe that the detection and segmentation of objects pose a significant challenge in our proposed dataset. Image samples for every traffic intersection in our dataset are shown in Figure 4.

TABLE X
KEY CHARACTERISTICS OF THE AAU RAINSNOW AND THE BADWEATHER (BW) CHANGEDETECTION.NET DATASETS. THE APPROXIMATE DURATION OF THE BADWEATHER SEQUENCES ARE CALCULATED WITH 20 FRAMES/SECOND. AN 'L' IN RAIN OR THUNDERSTORM INDICATES LIGHT RAIN AND LIGHT THUNDERSTORMS, RESPECTIVELY

| Sequence | Time of day (24 H) | Duration (minutes) | Average number of objects per frame | Average object size (pixels) | Rain | Snow (X), (F)og | Thunderstorm | Estimated temp. (°C) |
|---|---|---|---|---|---|---|---|---|
| Egensevej-1 | 18:00 | 5.0 | 3.10 | 636 | L | | | 5 |
| Egensevej-2 | 13:03 | 5.0 | 3.21 | 304 | L | X | | 2 |
| Egensevej-3 | 17:00 | 5.0 | 4.43 | 409 | X | | | 2 |
| Egensevej-4 | 18:00 | 5.0 | 4.83 | 448 | | X | | 1 |
| Egensevej-5 | 03:00 | 5.0 | 0.28 | 687 | X | | | 3 |
| Hadsundvej-1 | 20:06 | 4.0 | 6.69 | 795 | | | X | 19 |
| Hadsundvej-2 | 15:23 | 5.0 | 16.8 | 1293 | X | | L | 13 |
| Hjorringvej-1 | 06:04 | 5.0 | 3.48 | 1451 | X | | | 12 |
| Hjorringvej-2 | 16:00 | 5.0 | 13.7 | 1589 | L | | | 12 |
| Hjorringvej-3 | 19:12 | 5.0 | 6.62 | 1317 | X | | | 11 |
| Hjorringvej-4 | 07:00 | 5.0 | 6.49 | 1926 | L | | | 8 |
| Hobrovej-1 | 05:00 | 5.0 | 1.63 | 2438 | L | | | 14 |
| Ringvej-1 | 20:05 | 5.0 | 3.5 | 1033 | L | | | 12 |
| Ringvej-2 | 05:05 | 4.0 | 0.86 | 4533 | | F | | 5 |
| Ringvej-3 | 17:54 | 5.0 | 3.88 | 983 | L | | | 14 |
| Hasserisvej-1 | 13:12 | 5.0 | 4.85 | 560 | X | | | 11 |
| Hasserisvej-2 | 10:00 | 5.0 | 5.75 | 703 | L | | | 19 |
| Hasserisvej-3 | 13:00 | 5.0 | 7.58 | 616 | L | | | 19 |
| Ostre-1 | 06:10 | 5.0 | 2.50 | 771 | X | | | 15 |
| Ostre-2 | 10:00 | 5.0 | 10.7 | 896 | L | | | 10 |
| Ostre-3 | 18:00 | 5.0 | 4.98 | 514 | X | | X | 11 |
| Ostre-4 | 19:20 | 5.0 | 3.10 | 672 | X | | | 10 |
| BW/blizzard | - | 5.8 | 0.73 | 2391 | | X | | - |
| BW/skating | - | 3.3 | 0.58 | 6522 | | X | | |
| BW/snowFall | - | 5.4 | 0.15 | 7803 | | X | | - |
| BW/wetSnow | - | 2.9 | 0.53 | 3261 | | X | | - |

## A. Annotations

All frames of the BadWeather sequences are annotated at pixel level by the ChangeDetection.net initiative. For our dataset, the manual pixel-level annotation is complicated by the smaller size of apparent objects and many reflections. Thus, in order to make the annotations feasible, we have randomly selected 100 frames for each sequence from a uniform distribution. In a five-minute sequence at 20 frames per second, this means that, on average, an annotated frame is available every three seconds. We believe that the strong correlation between subsequent frames and the smooth motion of the road users enables us to achieve a good approximation of the entire sequence by annotating only a small subset of the frames. Consequently, we would rather spend time annotating more scenes that can capture a variety of challenging weather conditions than annotating a single sequence in its entirety. The annotation of our dataset is usually performed on the RGB images and mapped to the thermal images via a planar homography. In the case of severe reflections, the thermal image is used to guide the annnotations instead. We use the AAU VAP Multimodal Pixel Annotator [75] for drawing the annotations. We have marked 'do not care' care zones in areas without road users and when all objects are very small in a particular region, for instance the top of the surveillance video. Examples hereof are shown in Figure 5b.

[7] https://www.wunderground.com/

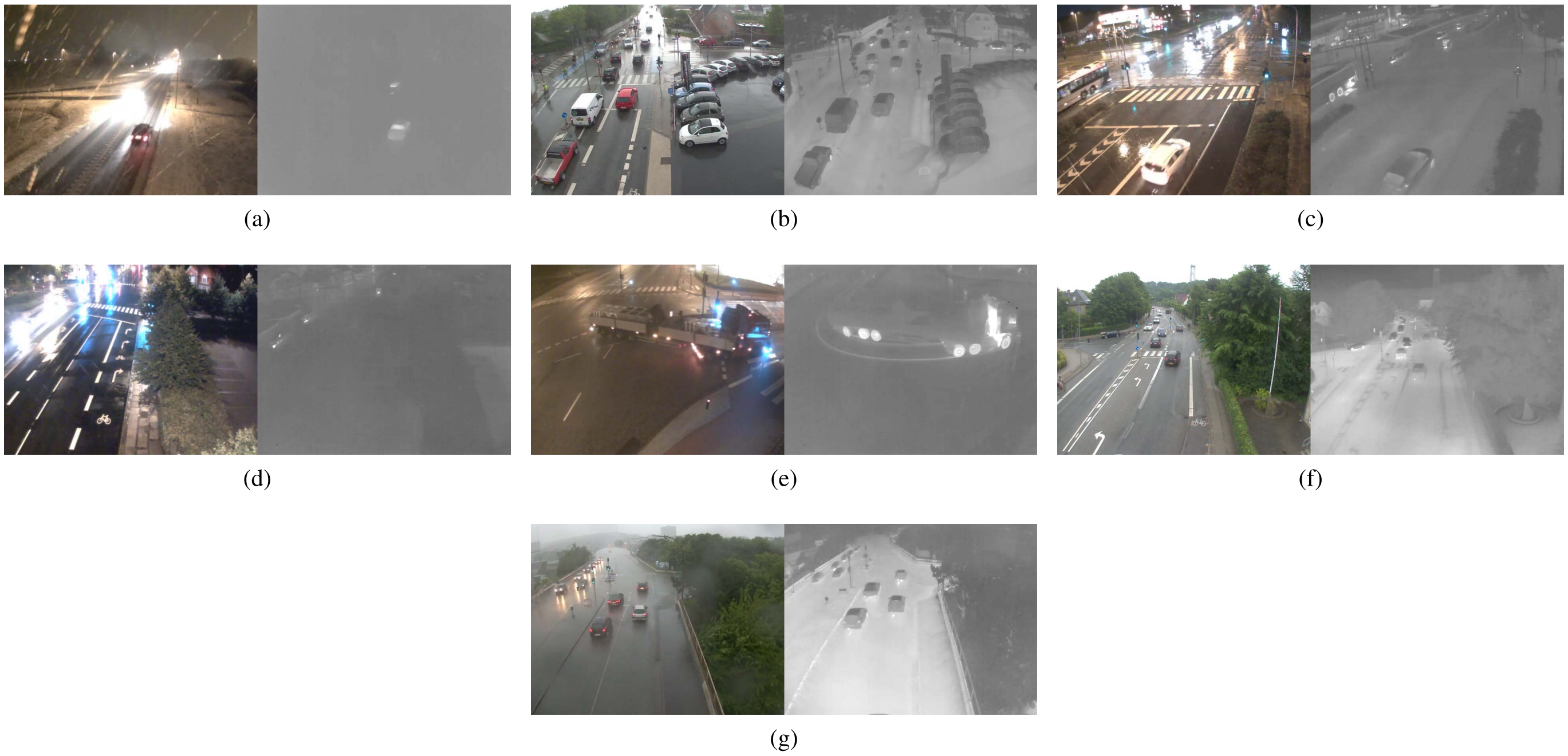

Fig. 4. Samples of each of the seven traffic intersections of the AAU RainSnow dataset. One sample is shown for each sequence with corresponding RGB and thermal image. For improved visibility, contrast is adjusted for all thermal images, except for the Hadsundvej sequence. (a) Egensevej. (b) Hadsundvej. (c) Hjorringvej. (d) Hobrovej. (e) Ringvej. (f) Hasserisvej. (g) Ostre.

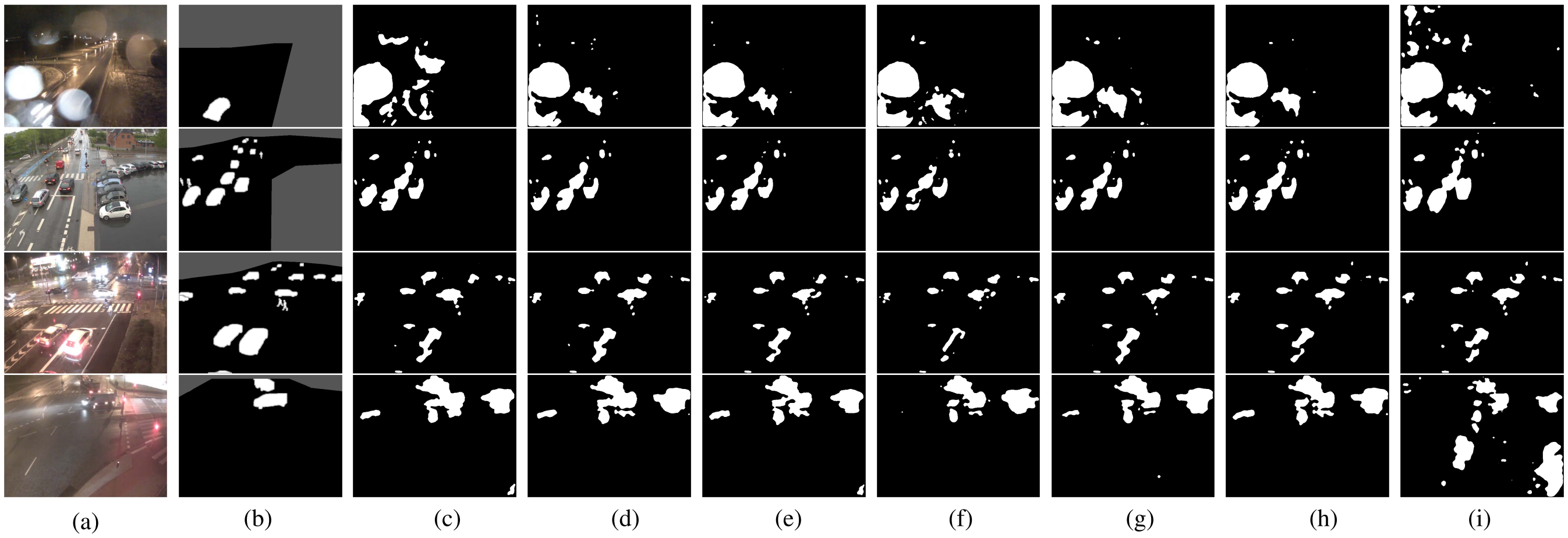

Fig. 5. Segmentation results on the AAU RainSnow dataset by the SuBSENSE algorithm [11]. Each row represents the results of different rain removal algorithms on a single frame. Sequences from top to bottom: Egensevej-5, Hadsundvej-1, Hjorringvej-4, and Ringvej-2. Gray areas indicate don't care zones. (a) RGB. (b) Ground truth. (c) Original. (d) Median. (e) Garg2007. (f) Kang2012. (g) Kim2015. (h) Fu2017. (i) Zhang2017.

## V. Evaluation Protocol

We will evaluate whether the rain removal algorithms introduced in Section III make a difference when used in a traditional computer vision pipeline that includes segmentation, tracking, and instance segmentation. In other words, a successful rain removal algorithm should improve the ability of subsequent algorithms to segment objects and perform feature tracking. In the context of traffic surveillance, the objects are road users. As such, the visual quality of the rain removed images or videos is not a concern as long as the subsequent traffic surveillance algorithms improves. If one wants to assess the visual quality of the rain removed images, the mean opinion score from multiple human assessments could be used. A user study is conducted in [29] to consider the most favorable result of several rain removal algorithms.

We use the AAU RainSnow dataset and the BadWeather sequences described in Section IV as the evaluation dataset. In order to run the rain removal algorithms on the datasets, the implementation should be available. However, for most

TABLE XI

EVALUATED RAIN REMOVAL ALGORITHMS. THE PROCESSING TIME PER IMAGE IS MEASURED ON AN INTEL CORE I7-3770 CPU WITH NVIDIA 1080TI GRAPHICS. *THE METHOD FROM ZHANG2017 IS GPU-BOUND, ALL OTHER METHODS ARE CPU-BOUND

| Paper | AuthorYear | Input | Main detection method | Learning | Processing time per image |
|---|---|---|---|---|---|
| [3] | Zhang2017 | Image | Conditional Generative Adversarial Network (DCGAN) | Offline | 0.40 s* |
| [24] | Fu2017 | Image | Convolutional Neural Network | Offline | 0.69 s |
| [20] | Kim2015 | Video | Sparse dictionary | Offline | 31.12 s |
| [43] | Kang2012 | Image | Sparse dictionary | Online | 78.72 s |
| [1] | Garg2007 | Video | Photometric constraint, streak orientation, size | Fixed | 0.81 s |
| - | 'Median' | Image | Spatial $3 \times 3$ pixels mean filter | Fixes | 0.06 s |

of the algorithms listed in Table IV, the implementation is not publicly available. Additionally, in most cases, it is not possible to re-implement the algorithms due to missing details in the original papers. Fortunately, the implementations of:

- Zhang *et al.* [3]
- Fu *et al.* [24]
- Kim *et al.* [20]
- Kang *et al.* [43]

are publicly available and will be used for comparison. Furthermore, we have implemented the rain removal algorithm by Garg and Nayar [1] as their work is generally considered the cornerstone from which many video-based rain removal algorithms are built. Our implementation is publicly available[8] and also provides links to the implementations listed above.

As the baseline for rain removal, we have added a spatial $3 \times 3$ pixels mean filter, which makes the image more smooth and may reduce the amount of rain. The evaluated rain removal algorithms are listed in Table XI. We use the default parameters from the original papers and list the average image processing time for every algorithm.

It should be noted that, although the dataset consists of video sequences, we have included both single-image and video-based rain removal algorithms. Although it is the general impression that video-based rain removal is significantly easier than single-image based rain removal, it has not been experimentally verified whether video-based algorithms use this advantage to outperform single-image based methods. Thus, we would like to find out by including algorithms from both categories.

The rain removal algorithms of Table XI undergo a two-tied evaluation on a segmentation, instance segmentation, and feature tracking pipeline. In the following, we will describe the protocol of the three evaluation pipelines.

### A. Segmentation

We evaluate the performance of rain removal algorithms under a traditional segmentation pipeline by running the rain removal algorithms in a separate pass and then running the segmentation algorithms on top of the rain-removed imagery. In order to select a segmentation algorithm that is representative of the state-of-the-art, we look to the results of the ChangeDetection.net challenge [6]. Although recent advantages in convolutional neural networks have led to superior performance of supervised segmentation methods as seen in Table III, we turn to the unsupervised methods instead. We believe that, in order for a segmentation method to be applicable in a real-world traffic surveillance context, the method should work out-of-the-box for non-experts and not require hand tuning in the form of parameters or training samples.

A representative of a top 3 unsupervised segmentation method is the 'SuBSENSE' algorithm [11]. SuBSENSE is a method that builds on the spatial diffusion step introduced in ViBE [76]. Instead of relying only on color information for the pixel description, SuBSENSE includes information of the local neighbors by computing Local Binary Similarity Patterns (LBSP) for every pixel. Based on a majority vote of the LBSP and local pixel values, the pixel is classified as either foreground or background.

Furthermore, we include the Mixture of Gaussians (MoG) method as modified by Zivkovic [77] as this is a classic segmentation method that is well understood and often used as a baseline for comparisons.

We use the F-measure to measure and compare the performance of the rain removal algorithms in the segmentation context. The F-measure is a widely used metric for evaluating change detection algorithms and has been found to agree well with the overall ranking computed from several metrics [6]. The F-measure is computed as:

$$F = 2 \cdot \frac{\mathrm{Pr} \cdot \mathrm{Re}}{\mathrm{Pr} + \mathrm{Re}} \tag{13}$$

where Pr (precision) is defined as:

$$\mathrm{Pr} = \frac{\mathrm{TP}}{\mathrm{TP} + \mathrm{FP}} \tag{14}$$

and Re (recall) is defined as:

$$\mathrm{Re} = \frac{\mathrm{TP}}{\mathrm{TP} + \mathrm{FN}} \tag{15}$$

where TP is the number of true positives, FP is the number of false positives, and FN is the number of false negative classified pixels in a sequence. The balance between recall and precision might be fine-tuned by adjusting the intrinsic parameters of the segmentation methods. For these experiments, we use the settings of the original authors. The evaluation of segmentation is performed as follows:

**for** Every video sequence in Table X **do**
  Run the segmentation algorithms on the unmodified, original frames of the video
  Compute the F-measure of the segmented frame and the ground truth
  **for** Every rain removal algorithm in Table XI **do**

[8] https://bitbucket.org/aauvap/rainremoval

Run the rain removal algorithm on each frame and save the rain-removed frame
Run the segmentation algorithms on the rain-removed frames and save the result
Compute the F-measure of the segmented frame and the ground truth
**end for**
**end for**

The results hereof are presented in Section VI-A.

### B. Instance Segmentation

The traditional segmentation challenge handles the separation of foreground objects from the background. In instance segmentation, one also needs to differentiate between every single instance of an object and assign the correct class label.

For evaluation, we follow the philosophy that algorithms should run out-of-the-box with no fine-tuning on our dataset. We use the popular Microsoft COCO API [78] and report results in the main COCO competition metric, average precision (AP) measured over intersection over union ratios in the range from 0.50 to 0.95 with intervals of 0.05. We have selected two instance segmentation algorithms for evaluation: Fully Convolutional Instance Segmentation (FCIS) [79] which won the 2016 COCO segmentation challenge and Mask R-CNN [80] which outperformed the FCIS network and ranked 3rd in the 2017 COCO segmentation challenge. Both algorithms are trained on the ImageNet [81] and COCO [78] datasets. Experimental results showed that the assigned class labels from both instance segmentation algorithms did not agree well with the ground truth of our RainSnow dataset. For instance, most cars were classified as trucks. Therefore, we decided to measure the precision of the class-agnostic instance segmentation by setting the 'useCats' parameter of the COCO API to 'false'. The evaluation is only performed on the AAU RainSnow dataset, as the ChangeDetection.net dataset is incompatible with the COCO evaluation format.

### C. Feature Tracking

We adopt the forward-backward feature-point tracking method used by Barnum *et al.* [19]. For every $n$ frames in a sequence, we select the 200 strongest features [82] and track them in the next $m$ frames using the Lucas-Kanade tracker [83]. After $m$ frames, the features are tracked when the sequence is played backwards to the point in time where the features were instantiated. We calculate the tracking accuracy by measuring the distance between the start and end positions of the tracked features. Similar to [19], we report the number of successfully tracked features within an error margin of 1 and 5 pixels.

Inspired by Barnum *et al.* [19], we have chosen the forward-backward tracking point accuracy for the following reasons:

- The tracking accuracy is correlated to the ability of the rain removal algorithms to preserve non-rain high-frequency components.
- The measure does not require ground truth and thus scales with the length of the sequence.
- If the tracking of a feature point is confused by the spatio-temporal fluctuations of a rain streak, the tracking accuracy should improve on rain-removed imagery.

Barnum *et al.* [19] evaluate the tracking accuracy once on the entire sequence, i.e. $n = l$ and $m = l$, where $l$ is the length of the video sequence, approximately five seconds. As our sequences are much longer, we need many separate instances of the forward-backward feature tracking. We have empirically found $n = 1.5$ s and $m = 12$ s, meaning that for every 1.5 seconds, we select the 200 strongest features which are tracked for 12 seconds forwards, then backwards. In our experience, different values of $n$ and $m$ only change the magnitude of the results. The results of the feature tracking are presented in Section VI-C.

## VI. Results

As described in Section V, we evaluate the rain removal algorithms of Table XI with respect to the performance of segmentation, instance segmentation, and feature tracking algorithms on the rain-removed sequences of Table X.

### A. Segmentation

The segmentation results, as indicated by the F-measure for the Mixture of Gaussians (MoG) and the SuBSENSE (SuB) segmentation algorithms, are listed in Table XII and plotted in Figure 6b. If we take a look at the segmentation results on the unmodified video, i.e. the non-rain-removed frames, we may note that he proposed AAU RainSnow dataset imposes a significant challenge to segmentation algorithms. The F-measure of our dataset varies from 0.11 to 0.66, even for the state-of-the-art SubSENSE method. The MoG method fare even worse, with F-measures in the range from 0.13 to 0.34. Segmentation results are much better on the BadWeather sequences, where the F-measure is in the range of 0.80 to 0.89 for the SuBSENSE method.

When looking at the segmentation results of the rain-removed images, we should take note of the aforementioned differences in segmentation performance and the inherent differences between the AAU RainSnow and BadWeather datasets as described in Section IV. On the AAU RainSnow dataset, we see from Table XII that the GAN-based convolutional neural network by Zhang *et al.* [3] gives an average increase of 28.5% in the segmentation performance of the SuBSENSE algorithm, whereas the same algorithm results in an average decrease of 38.6% on the BadWeather sequences. Except for the combination of MoG on the rain-removed videos in the method by Kim *et al.* [20], all rain removal algorithms reduce the performance of segmentation algorithms on the BadWeather dataset. Nevertheless, all rain removal algorithms give a performance increase when using the SuBSSENSE method based on the AAU RainSnow dataset. Examples of the visual segmentation results on the AAU Rainsnow database are shown in Figure 5.

It is difficult to give an unequivocal explanation of the cause of the great difference seen in results between the two datasets. This variance may be caused by a combination several factors:

TABLE XII

EVALUATION OF SEGMENTATION PERFORMANCE ON EACH SEQUENCE. THE ABSOLUTE F-MEASURE IS REPORTED FOR THE ORIGINAL, NON-RAIN-REMOVED FRAMES. OTHER RESULTS ARE RELATIVE TO THE ORIGINAL RESULTS OF EACH SEQUENCE, IN PERCENTAGES. MOG: MIXTURE OF GAUSSIANS [77]. SUB: SUBSENSE [11]. BEST RESULT OF A SEQUENCE IS INDICATED IN BOLD. CATEGORY AVERAGES ARE COMPUTED FROM THE SUM OF ABSOLUTE F-MEASURES

| Sequence | Original | | Median | | Garg2007 [1] | | Kang2012 [43] | | Kim2015 [20] | | Fu2017 [24] | | Zhang2017 [3] | |
|---|---|---|---|---|---|---|---|---|---|---|---|---|---|---|
| | MoG | SuB | MoG | SuB | MoG | SuB | MoG | SuB | MoG | SuB | MoG | SuB | MoG | SuB |
| Egensevej-1-5 | 0.13 | 0.11 | -5.77 | 13.9 | -1.50 | 26.2 | -21.4 | 12.0 | **8.69** | 17.4 | -0.29 | 18.4 | -4.20 | **37.9** |
| Hadsundvej-1-2 | 0.19 | 0.66 | -10.8 | 4.48 | -7.67 | 2.50 | -27.1 | -9.46 | 1.41 | 3.78 | -4.15 | 6.92 | **14.8** | **21.4** |
| Hasserisvej-1-3 | 0.34 | 0.61 | -4.16 | 10.6 | -1.41 | 12.7 | -4.38 | 11.4 | -4.17 | 12.1 | 0.44 | 18.6 | **10.7** | **24.4** |
| Hjorringvej-1-4 | 0.21 | 0.52 | -1.57 | 7.73 | -1.57 | 6.19 | -20.2 | 5.48 | -1.58 | 6.64 | 1.86 | 9.54 | **14.7** | **22.1** |
| Hobrovej-1 | 0.28 | 0.36 | 1.61 | -1.93 | 4.52 | 9.02 | -36.8 | **42.7** | 8.75 | 21.8 | **10.8** | 0.63 | 3.93 | -23.8 |
| Ostre-1-4 | 0.26 | 0.49 | 10.5 | 10.9 | 10.8 | 13.2 | -27.4 | -9.87 | 11.3 | 12.3 | 14.6 | 14.8 | **28.4** | **23.2** |
| Ringvej-1-3 | 0.23 | 0.19 | 16.1 | 5.40 | 14.6 | 13.9 | -0.39 | **76.0** | 10.7 | 25.0 | 16.9 | 8.89 | **23.7** | 5.51 |
| BadWeather/blizzard | 0.25 | 0.85 | -98.6 | -99.7 | -99.0 | -99.7 | -90.5 | -91.1 | -5.03 | -1.43 | -7.17 | -1.47 | **15.5** | **0.53** |
| BadWeather/skating | 0.28 | 0.89 | -3.57 | -7.39 | 7.33 | -11.9 | -76.7 | -87.2 | **9.22** | **-7.30** | -6.84 | -7.75 | -79.3 | -91.4 |
| BadWeather/snowFall | 0.18 | 0.88 | 0.57 | -18.3 | 5.87 | -18.1 | -74.8 | -90.1 | 32.4 | **-15.8** | 12.2 | -17.3 | **38.2** | -18.9 |
| BadWeather/wetSnow | 0.25 | 0.80 | -0.23 | -20.3 | 2.96 | -26.3 | -87.7 | -95.3 | **16.7** | -19.9 | 5.11 | **-18.5** | -37.4 | -47.9 |
| AAU RainSnow avg. | 0.22 | 0.40 | -0.16 | 6.36 | 1.31 | 9.28 | -18.6 | 10.5 | 4.55 | 3.42 | 4.29 | 10.5 | **14.0** | **28.5** |
| BadWeather avg. | 0.25 | 0.87 | -31.5 | -37.0 | -24.4 | -39.4 | -82.4 | -89.5 | **9.88** | **-8.75** | -2.33 | -9.15 | -18.7 | -38.6 |

- The segmentation of the BadWeather sequences already produced good results, rendering it difficult to improve.
- The average object size of the BadWeather sequences is 3.5 times larger than the average object size of the AAU RainSnow dataset. This difference may occur because the segmentation of smaller objects benefits from the removal of rain streaks, whereas the segmentation of larger objects is more resilient to the fluctuations from rain and snow. In this case, the spatio-temporal low-pass filtering of the rain removal algorithms may eventually harm the segmentation performance.

It should be noted that further experimentation on other datasets is needed in order to fully understand the underlying causes.

The visual results of Figure 5 based on the AAU RainSnow dataset confirm that raindrops on the lens and reflections on the road pose a challenge to the segmentation process. However, even under these challenging conditions, the results from Table XII show that the evaluated rain removal algorithms improve the segmentation. Two notable exceptions are the Egensevej-5 and Ringvej-2 sequences, which are shown in the top and bottom rows of Figure 5, respectively. On the Egensevej-5 sequence, the best rain removal algorithm decreases the segmentation performance of the SuBSENSE algorithm by 44%, whereas the remaining algorithms perform even worse. On the Ringvej-2 sequence, the otherwise top performing rain removal methods of Fu *et al.* [24] and Zhang *et al.* [3] fail to improve the segmentation results at all. One possible explanation could be that the segmentation performance of the two original sequences is quite poor, with F-measures of 0.05 and 0.14 for the SuBSENSE method on the Egensevej-5 and Ringvej-2 sequences, respectively. If the underlying phenomena responsible for the degradation of the visual quality are not related to rainfall and snowfall, the corrections from rain removal algorithms may be ill-behaved and degrade the results.

However, the remaining sequences of the AAU RainSnow dataset show decent increases in segmentation performance for most rain removal algorithms, even for sequences in which the segmentation is relatively hard. The improvement in segmentation results from relatively 'good' sequences with good illumination and few shadows, such as the Hadsundvej sequences, indicates that rain removal algorithms could be a suitable preprocessing step for traffic surveillance scenes under rain and snow when improved performance of subsequent traditional segmentation algorithms is required.

### B. Instance Segmentation

The average precision of the instance segmentation methods Mask R-CNN [80] and FCIS [79] are shown in Table XIII and visualized with box plots in Figure 6c. It is evident from the results of the original sequences that the Mask R-CNN method outperforms the FCIS method by a large margin, resulting in a AP of 0.33 and 0.07 on the entire AAU RainSnow dataset, respectively. If we compare the instance segmentation results with the traditional segmentation results of Table XII, both segmentation approaches struggle with the Egensevej sequences. On the Hobrovej sequence, the traditional segmentation methods fares well whereas the instance segmentation methods breaks down. One should note, however, that the instance segmentation methods does not take temporal information into account, which makes the segmentation increasingly harder under difficult weather.

All evaluated rain removal algorithms fail to improve the instance segmentation results of the Mask R-CNN. The best performing algorithm of Zhang *et al.* [3] degrades the AP by 3.45% while the worst performing method by Kang *et al.* [43] degrades the result by 36.3%.

On the contrary, all rain removal algorithms but the method by Kang *et al.* [43] improves the instance segmentation results of the FCIS method. The best rain removal algorithm on the FCIS method is the 3x3 spatial mean filter and the CNN-based method by Zhang *et al.* [3] which improves the result by 27.6% and 25.2%, respectively. However, even with these improvements, the FCIS method is inferior to Mask R-CNN.

It is remarkable that the simple median filter outperforms the dedicated rain removal algorithms with the FCIS method and lies close to other algorithms with the Mask R-CNN.

TABLE XIII

EVALUATION OF INSTANCE SEGMENTATION PERFORMANCE ON EACH SEQUENCE. THE AVERAGE PRECISION (AP[.5:.05:.95] IS REPORTED FOR THE ORIGINAL, NON-RAIN-REMOVED FRAMES. OTHER RESULTS ARE RELATIVE TO THE ORIGINAL RESULTS OF EACH SEQUENCE, IN PERCENTAGES. FB: MASK R-CNN FROM FACEBOOK [80]. MS: FCIS FROM MICROSOFT [79]. BEST RESULT OF A SEQUENCE IS INDICATED IN BOLD

| Sequence | Original | | Median | | Garg2007 [1] | | Kang2012 [43] | | Kim2015 [20] | | Fu2017 [24] | | Zhang2017 [3] | |
|---|---|---|---|---|---|---|---|---|---|---|---|---|---|---|
| | FB | MS | FB | MS | FB | MS | FB | MS | FB | MS | FB | MS | FB | MS |
| Egensevej-1-5 | 0.10 | 0.02 | -9.32 | 2.30 | **-7.27** | -23.3 | -49.3 | -8.33 | -10.1 | -6.97 | -28.4 | -55.8 | -23.2 | **4.86** |
| Hadsundvej-1-2 | 0.45 | 0.07 | -4.99 | **95.3** | -3.93 | 86.8 | -26.6 | 50.4 | -3.41 | 92.7 | -29.0 | 58.9 | **1.76** | 89.3 |
| Hasserisvej-1-3 | 0.44 | 0.12 | -6.91 | 1.48 | -5.76 | -0.95 | -9.70 | 0.31 | -5.46 | 0.71 | -0.04 | 4.18 | **0.88** | **5.43** |
| Hjorringvej-4-1 | 0.34 | 0.10 | -6.28 | -0.38 | -7.27 | -4.03 | -38.3 | -15.7 | -6.76 | **-0.04** | -44.2 | -26.1 | **-6.03** | -1.79 |
| Hobrovej-1 | 0.00 | 0.00 | -49.9 | 17.1 | **219.1** | **342.5** | -79.6 | -15.2 | -60.9 | 65.8 | -82.1 | -31.9 | -27.5 | -87.4 |
| Ostre-1-4 | 0.29 | 0.07 | -14.7 | 7.51 | -4.41 | 4.20 | -68.2 | -60.4 | -4.67 | 3.64 | -6.51 | 6.47 | **-2.19** | **15.5** |
| Ringvej-1-3 | 0.27 | 0.05 | -8.52 | 3.47 | -12.6 | -13.5 | -53.5 | -56.5 | -10.5 | -5.24 | **-3.63** | **17.5** | -25.5 | -27.4 |
| AAU RainSnow avg. | 0.33 | 0.07 | -7.70 | **27.6** | -5.89 | 21.7 | -36.3 | -1.83 | -5.77 | 25.2 | -24.5 | 11.4 | **-3.45** | 24.9 |

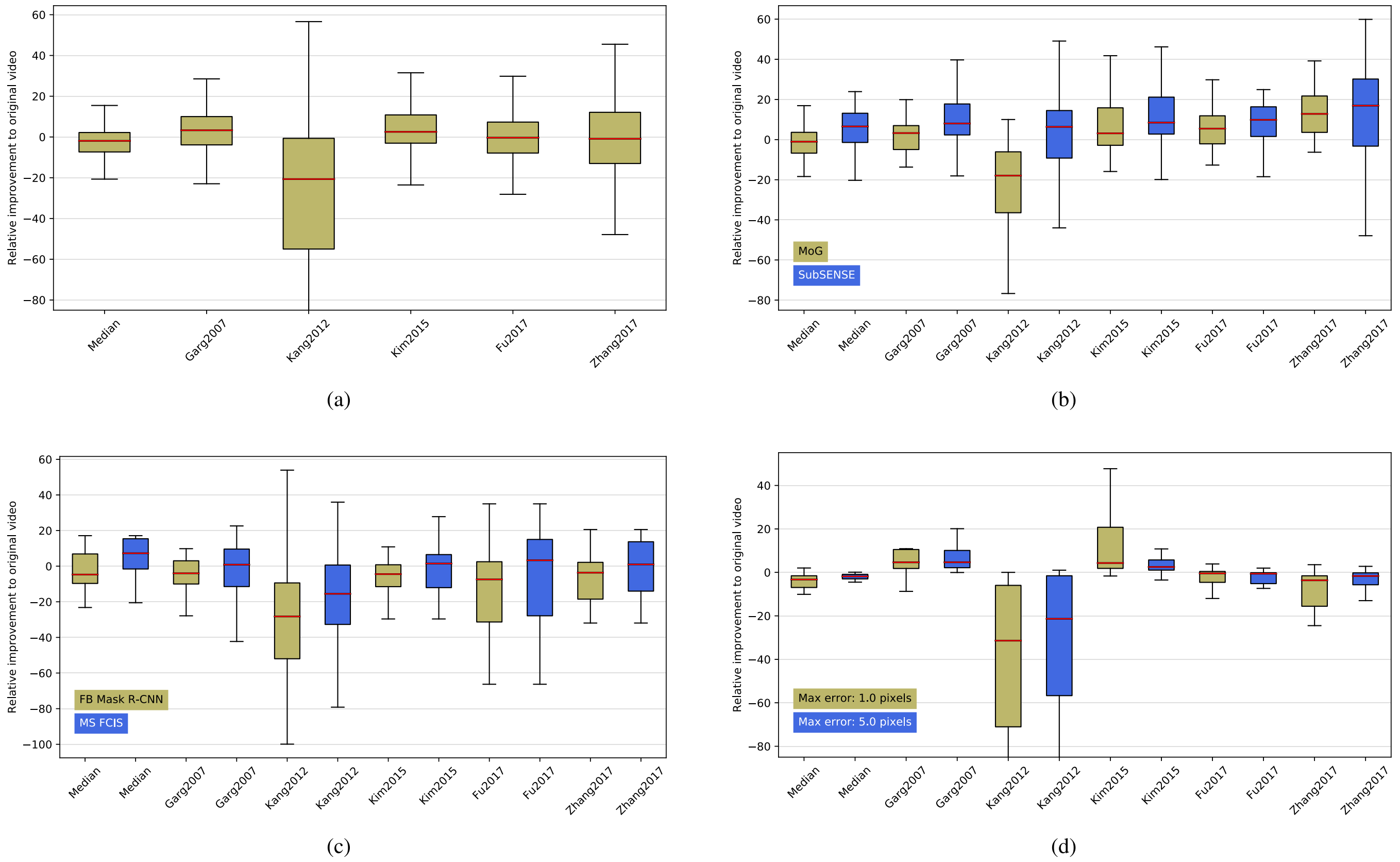

Fig. 6. Box and whiskers plot of the relative improvement as a result of the pre-processing by rain removal algorithms. (a) Overall. (b) Segmentation. (c) Instance segmentation. (d) Feature tracking.

Both instance segmentation methods have been trained on the ImageNet and COCO datasets and are thus designed to respond to images that resemble these training sets. Our AAU RainSnow dataset is a different, surveillance-type dataset with many small objects that does not necessarily resemble these training datasets. Given a dissimilar dataset, the noise and alterations by the applied rain removal algorithms might push the images out of the visual manifold that the instance segmentation methods have been trained on.

### *C. Feature Tracking*

The results of the forward-backward feature point tracking are shown in Table XIV and the box plots of Figure 6d. If we look at the average results on both datasets, it is observed that the rain removal algorithm by Zhang *et al.* [3], which was superior when evaluated on the segmentation pipeline, consistently deteriorates the feature tracking performance. It should be noted that the algorithm by Zhang *et al.* [3] is a single-frame based method and does not incorporate the temporal information when removing the rain. In fact, all the evaluated single-frame rain removal algorithms deteriorate the feature-tracking results (Median, Kang *et al.* [43], Fu *et al.* [24], Zhang *et al.* [3]).

If we look at the results of the video-based rain removal methods by Garg and Nayar [1] and Kim *et al.* [20], we observe a general increase in feature tracking performance. The relatively simple method by Garg and Nayar [1] contributes to an average increase in the number of successfully

TABLE XIV

EVALUATION OF FORWARD-BACKWARD FEATURE TRACKING ON EACH SEQUENCE. THE NUMBER OF SUCCESSFULLY TRACKED FEATURES WITH AN ERROR MARGIN OF 1.0 AND 5.0 PIXELS IS REPORTED FOR THE ORIGINAL, NON-RAIN-REMOVED FRAMES. OTHER RESULTS ARE RELATIVE TO THE ORIGINAL RESULTS OF EACH SEQUENCE, IN PERCENTAGES. BEST RESULT OF A SEQUENCE IS INDICATED IN BOLD. CATEGORY AVERAGES ARE COMPUTED FROM THE SUM OF TRACKED FEATURES

| Sequence | Original | | Median | | Garg2007 [1] | | Kang2012 [43] | | Kim2015 [20] | | Fu2017 [24] | | Zhang2017 [3] | |
|---|---|---|---|---|---|---|---|---|---|---|---|---|---|---|
| | 1.0 | 5.0 | 1.0 | 5.0 | 1.0 | 5.0 | 1.0 | 5.0 | 1.0 | 5.0 | 1.0 | 5.0 | 1.0 | 5.0 |
| Egensevej-1-4 | 44601 | 64486 | -5.51 | -3.31 | 21.6 | 22.9 | -68.2 | -57.8 | **52.5** | **32.1** | -3.20 | -5.76 | -27.6 | -21.6 |
| Hadsundvej-1-2 | 46558 | 54592 | -3.17 | -1.84 | **5.07** | **4.63** | -7.95 | -3.61 | 3.39 | 2.25 | 0.12 | -0.01 | -1.33 | -0.40 |
| Hasserisvej-1-3 | 88155 | 94372 | -2.52 | -0.81 | **2.19** | **2.73** | -0.40 | 0.08 | 1.52 | 1.88 | -1.53 | -0.85 | -3.08 | -1.84 |
| Hjorringvej-1-4 | 87324 | 104680 | -2.88 | -1.90 | **5.74** | **7.52** | -7.42 | -2.21 | 5.20 | 4.68 | 0.19 | -0.35 | 0.25 | 0.86 |
| Hobrovej-1 | 32364 | 35502 | -1.53 | -0.62 | **5.00** | **1.59** | -19.6 | -9.67 | 4.07 | 1.03 | 0.26 | -0.11 | -4.53 | -0.35 |
| Ostre-1-4 | 106934 | 129240 | -2.91 | -1.77 | **5.21** | **3.21** | -55.1 | -33.5 | 2.72 | 0.27 | -7.37 | -6.98 | -2.48 | -0.81 |
| Ringvej-1-3 | 67809 | 75810 | -1.98 | -1.05 | 1.44 | 1.42 | -32.8 | -19.7 | **2.23** | **1.53** | -3.15 | -2.90 | -6.54 | -1.93 |
| BadWeather/blizzard | 7630 | 17996 | -77.1 | -48.7 | -8.73 | **15.0** | -85.7 | -60.8 | **111.2** | 5.95 | -25.5 | -18.4 | -58.7 | -25.7 |
| BadWeather/skating | 1934 | 12002 | -39.9 | -12.8 | 315.7 | **18.1** | -100.0 | -100.0 | **455.9** | 15.7 | 61.0 | 1.97 | -45.6 | -17.4 |
| BadWeather/snowFall | 1649 | 9983 | -20.7 | -15.8 | 276.6 | 98.0 | -100.0 | -100.0 | **1093.8** | **163.8** | -12.0 | -3.90 | -36.9 | -6.29 |
| BadWeather/wetSnow | 9571 | 15483 | 2.02 | -1.41 | 45.9 | **4.26** | -2.48 | -3.97 | **47.7** | 2.72 | 9.98 | -1.79 | -3.18 | -0.84 |
| AAU RainSnow avg. | 473745 | 558682 | -2.87 | -1.65 | 5.72 | **6.01** | -27.1 | -18.5 | **7.72** | 5.45 | -2.63 | -2.89 | -5.07 | -3.15 |
| BadWeather avg. | 20784 | 55464 | -32.7 | -21.8 | 69.3 | 27.6 | -49.8 | -60.5 | **192.0** | **35.6** | -0.06 | -6.75 | -30.2 | -13.5 |

tracked feature points with a margin of error of 1 pixel of 5.72% and 69.3% on the AAU RainSnow and BadWeather datasets, respectively. Comparatively, the rain removal algorithm by Kim *et al.* [20] results in a modest improvement of 7.72% on the AAU RainSnow dataset. On the BadWeather dataset, the improvement is more pronounced with a corresponding performance increase of 192%. If we look at the average processing times per image listed in Table XI, the method by Garg and Nayar [1] is 38 times faster than the method by Kim *et al.* [20] and should thus be preferred due to superior speed.

As opposed to the segmentation results, which did not agree on the AAU RainSnow and BadWeather datasets, the feature tracking results on the AAU RainSnow and BadWeather datasets differ only by an order of magnitude.

In general, the results indicate that feature-point tracking on traffic surveillance videos benefits from the spatial low-pass filtering of the video-based rain removal algorithms.

## VII. CONCLUSION

We have studied the effects of rain and snow in the context of traffic surveillance and reviewed single-frame and video-based algorithms that artificially remove rain and snow from images and video sequences. The study shows that most of these algorithms are evaluated on synthetic rain and short sequences with real rain and their behavior in a realistic traffic surveillance context are undefined and not experimentally validated. In order to investigate how they behave in the aforementioned context, we have presented the AAU RainSnow dataset that features traffic surveillance scenes captured under rainfall or snowfall and challenging illumination conditions. We have provided annotated ground truth for randomly selected image frames of these sequences in order to evaluate how the preprocessing of the input video by rain removal algorithms will affect the performance of subsequent segmentation, instance segmentation, and feature tracking algorithms.

Based on their dominance in the field and their public availability, we selected six rain removal algorithms for evaluation, two video-based methods and four single-frame based methods. The results presented in Table XII show that the single-frame based rain removal method of Zhang *et al.* [3] improves the segmentation by 19.7% on average on the AAU RainSnow dataset. However, it deteriorates the performance on the BadWeather sequences of the public ChangeDetection.net dataset [6] and is not successful on a classical feature tracking pipeline. As a result, we achieve lower accuracy on forward-backward feature tracking on the rain-removed frames by Zhang *et al.* [3] than running the feature tracking on the unmodified original input frames. On the contrary, all video-based rain removal algorithms consistently improve the feature tracking results on the AAU RainSnow and BadWeather datasets. We received mixed results from the evaluation of instance segmentation methods. On a state-of-the-art method, the pre-processing by the evaluated rain removal algorithms decreased the segmentation performance. However, with the exception of the rain removal algorithm of Kang *et al.* [43], all rain removal algorithms improved the performance on a slightly older, less capable instance segmentation method.

If we look at the overall improvement across the three evaluation metrics as shown in Figure 6a, we observe a large variability in the performance of the rain removal algorithms. The simplest method, the spatial median filter, shows the lowest variability whereas the method from Kang *et al.* [43] shows the greatest variability and worst performance with a median improvement of $-20\%$. The CNN-based methods of Fu *et al.* [24] and Zhang *et al.* [3] both show a median improvement around 0%, with lower variability of the former method. The video-based methods of Garg and Nayar [1] and Kim *et al.* [20] show similar performance with a median improvement at 3.3 and 2.5%, respectively. When considering the processing time required by the method of Kim *et al.* [20], the well-established method from Garg and Nayar [1] is considered to be the best general-purpose rain removal algorithm.

In this paper, we aimed to answer the initial research question: Does rain removal in traffic surveillance matter? We must conclude that, as with other aspects of computer vision, this really depends on the application. Our experiments show that some applications benefit from rain removal, whereas other applications see their performance significantly reduced. Thus, rain removal algorithms should not be used as a general pre-processing tool in traffic surveillance, but they could be considered depending on the experimental results of the desired application. It should be noted that we have only tested the rain removal algorithms on video sequences in which it was actually raining or snowing. The behavior of these algorithms on non-rain sequences is still undefined. Further investigations could go into an intelligent switching system that enables such pre-processing systems based on the available contextual information.

In our experiments, we have chosen to evaluate the performance on three computer vision methods: segmentation, instance segmentation, and feature tracking. However, it is still an open question how rain removal algorithms perform when evaluated on other methods, such as classification, object tracking, and 3D reconstruction.

## Acknowledgment

This publication reflects only the author's views. The European Commission is not responsible for any use that may be made of the information it contains.

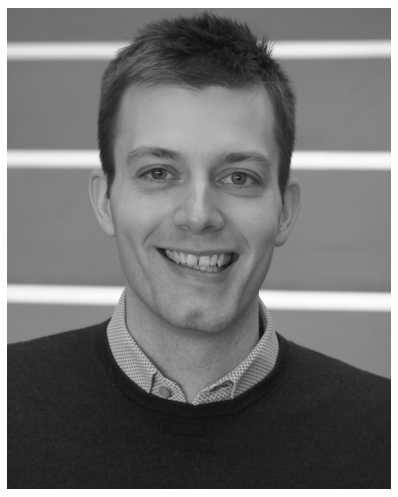

**Chris H. Bahnsen** received the B.Sc. degree in electrical engineering and the M.Sc. degree in vision, graphics, and interactive systems from Aalborg University, Aalborg, Denmark, in 2011 and 2013, respectively, where he is currently pursuing the Ph.D. degree in the field of visual analysis of people and traffic surveillance. He is a Visiting Scholar with the Advanced Driver Assistance Systems Research Group, Computer Vision Center, Universitat Autònoma de Barcelona, Spain. His main interests include computer vision and machine learning, particularly in the area of road traffic surveillance.

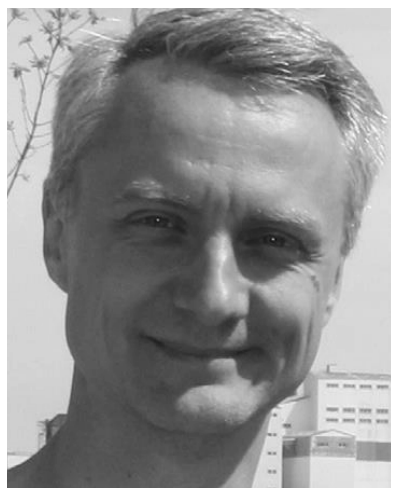

**Thomas B. Moeslund** received the M.Sc.E.E. and Ph.D. degrees from Aalborg University, Aalborg, Denmark, in 1996 and 2003, respectively. He is currently the Head of the Section for Media Technology and the Visual Analysis of People Laboratory, Aalborg University. His research interests include all aspects of computer vision, with a special focus on automatic analysis of people. He is involved in 35 (inter-) national research projects. He has authored over 250 peer-reviewed papers. He was a recipient of the Most Cited Paper Award in 2009, five best paper awards in 2010, 2013, 2016, and 2017, and the Teacher of the Year Award in 2010. He has (co-)chaired over 20 international workshops/tutorials. He serves as an associate editor for four international journals.